\documentclass[runningheads]{llncs}

\usepackage{eccv}

\usepackage{eccvabbrv}

\usepackage{graphicx}
\usepackage{booktabs}

\usepackage[accsupp]{axessibility}  

\def\eg{\emph{e.g}\onedot, }

\usepackage{hyperref}

\usepackage{orcidlink}
\usepackage{multirow}
\usepackage{makecell}
\usepackage{comment}
\usepackage{stackengine}
\usepackage{fontawesome5}
\usepackage{amsmath}
\usepackage{bbm}
\usepackage{kotex}
\usepackage{bbm}
\usepackage{enumitem}
\usepackage[table]{xcolor}
\usepackage{pifont}
\newcommand\blfootnote[1]{%
  \begingroup
  \renewcommand\thefootnote{}\footnote{#1}%
  \addtocounter{footnote}{-1}%
  \endgroup
}

\newcommand{\cmark}{\ding{51}}

\begin{document}

\title{What You Ask is What You Ground: \\Bridging Question Intent to Temporal Evidence for Grounded VideoQA} 

\titlerunning{Bridging Question Intent to Temporal Evidence for Grounded VideoQA}

\author{Jinhwan Seo\inst{1}\orcidlink{0000-0003-0495-7144} \and
Kyubeom Han\inst{1}\orcidlink{0009-0001-1196-4548} \and
Jumin Lee\inst{1}\orcidlink{0000-0002-0565-2679} \and
Junhyug Noh\inst{2}\textsuperscript{$\dagger$}\orcidlink{0000-0003-1239-8178} \and
Sung-eui Yoon\inst{1}\textsuperscript{$\dagger$}\orcidlink{0000-0002-7123-1119}
}

\authorrunning{J. Seo et al.}

\institute{KAIST \and Ewha Womans University\\
\email{\{jinhwan.seo,qbhan,jmlee\}@kaist.ac.kr, junhyug@ewha.ac.kr, sungeui@kaist.ac.kr}}

\maketitle
\blfootnote{\textsuperscript{$\dagger$}Corresponding authors.}
\begin{center}
    \captionsetup{type=figure}
    \includegraphics[width=0.99\linewidth]{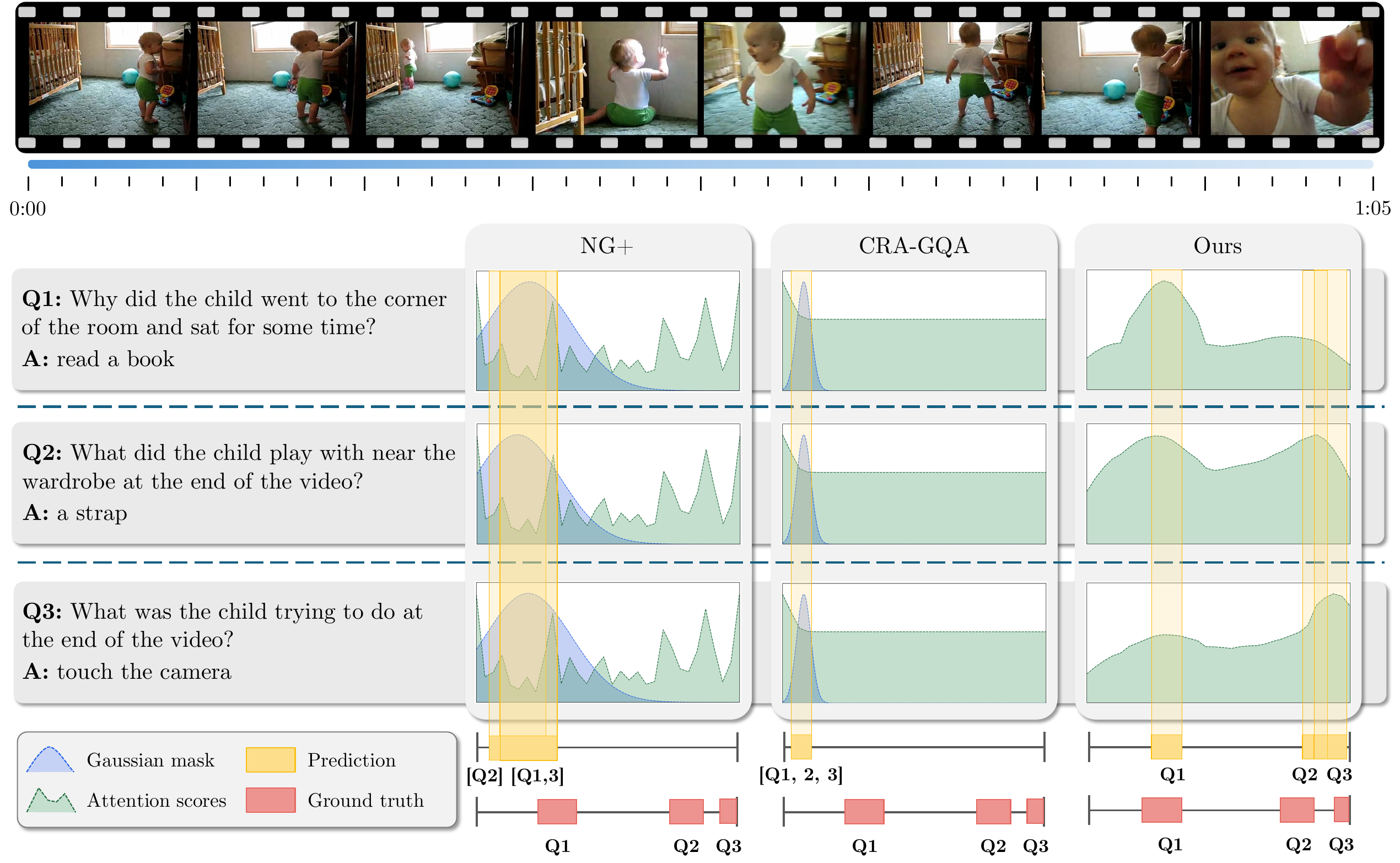}
    \captionof{figure}{Given a single video and three distinct questions targeting different temporal moments, prior methods yield nearly identical temporal predictions regardless of question intent. In contrast, GroundFormer produces three distinct segments, each aligned with the event required to reason toward the corresponding answer.}
    \label{fig:teaser}
\end{center}

\begin{abstract}
We study a critical yet overlooked failure mode in Grounded Video Question Answering: \emph{question-invariant grounding}, where models predict nearly identical temporal segments for different questions about the same video.
We trace this behavior to two structural limitations in prior common designs: (i) \emph{modality isolation} that fixes video representations before they receive question semantics, and (ii) \emph{weak question injection} inside the grounding module.
To address this, we propose GroundFormer, which conditions video features on question intent \emph{before} localization via learnable communication tokens that mediate directed visuo-lingual interaction.
On top of the question-conditioned features, a factorized MIL cross-attention couples answer selection with temporal evidence under candidate-level supervision, while Gaussian smoothing converts peaked attention into temporally coherent segments.
We further introduce a hierarchical multi-modal contrastive loss that aligns video, question, and answer embeddings across a two-pass training pipeline.
GroundFormer achieves state-of-the-art grounded VideoQA performance on NExT-GQA and STAR, substantially improving question-discriminative temporal grounding.

\end{abstract}
\section{Introduction}
\label{sec:intro}
Video Question Answering (VideoQA)~\cite{choi2021dramaqa,jang2017tgif,xiao2021next,antol2015vqa} has emerged as a primary benchmark for evaluating the multi-modal reasoning capabilities of vision-language models (VLMs)~\cite{radford2021learning,li2022blip,dai2023instructblip,liu2023visual,alayrac2022flamingo,wang2022internvideo}. Recent advances in large-scale pretraining of large language models~\cite{liu2019roberta,he2020deberta,devlin2019bert} have driven remarkable progress on the VideoQA task, yet the BlindQA experiment~\cite{xiao2024can} reveals that models without video input achieve performance on par with state-of-the-art methods, suggesting that predictions may reflect linguistic priors rather than visual understanding.
This experiment motivates Grounded VideoQA~\cite{xiao2024can}, which requires models to produce not only a correct answer but also the temporal segment from which it is derived, ensuring that answers are grounded in visual evidence. However, we identify a tendency that persists across existing methods~\cite{xiao2024can,liu2024timecraft,chen2025cross}: \emph{question-invariant grounding}, where models produce nearly identical temporal predictions for different questions within the same video.

As illustrated in~\cref{fig:teaser}, when multiple questions target different temporal events within the same video, existing methods such as NG+~\cite{xiao2024can} and CRA-GQA~\cite{chen2025cross} collapse all predictions onto the same segment regardless of question intent.
Each question asks about a distinct moment, yet the predicted grounding intervals are nearly indistinguishable across questions, revealing that previous methods~\cite{xiao2024can,chen2025cross} fail to condition on what each question asks.
We attribute such phenomena to two structural limitations: visual representations are fixed before any question signal reaches them, and the question is reduced to a single token at the grounding stage that is immediately absorbed by the dominant visual stream. As a result, the grounding module operates on representations that are blind to what each question asks.

In this work, we propose \textbf{GroundFormer}, a framework that addresses \emph{question-invariant grounding} by establishing a visuo-lingual communication pathway through which question intent is propagated into visual representations.
To counter modality isolation, we introduce learnable communication tokens that act as a bottleneck between the visual and linguistic branches: the tokens extract fine-grained question semantics from the linguistic stream and inject the captured intent into visual encoding.
To overcome the insufficient language signal in the grounding module, we formulate Grounded VideoQA as a Multiple Instance Learning (MIL) problem via cross-attention, where question-conditioned communication tokens serve as queries and video tokens as keys and values, such that temporal evidence is derived from question intent under candidate-level supervision alone.

In summary, our contributions are:
\begin{itemize}[
    topsep=0ex,          
]
    \item We introduce learnable \emph{communication tokens} as an explicit bottleneck between the visual and linguistic branches, enabling question intent to shape video representations \emph{before} grounding and alleviating modality isolation.
    \item We cast Grounded VideoQA as a \emph{multiple instance learning (MIL)} problem and propose a MIL cross-attention that couples answer prediction with temporal evidence, producing question-conditioned localization from candidate-level QA supervision alone.
    \item We achieve state-of-the-art Acc@GQA and grounding performance on NExT-GQA and STAR, and PIoU and GT-Pred Correlation analyses confirm that our grounding is more question-discriminative than prior methods.
\end{itemize}
\section{Related Work}
\label{sec:related}

\noindent\textbf{VideoQA.}
With the success of vision-language models (VLMs)~\cite{radford2021learning,li2022blip,li2023blip,dai2023instructblip,liu2023visual}, VideoQA has advanced from simple action recognition to tasks requiring reasoning over event sequences, causal relationships, and temporal dependencies across video~\cite{xiao2021next, tapaswi2016movieqa, choi2021dramaqa, jang2017tgif, min2024morevqa}. Early approaches employed RNNs and CNNs~\cite{jang2017tgif, sun2019learning, yin2019memory}, while recent methods adopted transformer-based architectures~\cite{sun2019learning, xiao2022video} and Q-Former frameworks~\cite{li2023blip,zhang2023video} that couple frozen visual encoders with large language models (LLMs) via learnable queries. Each generation improves QA capability, yet whether these gains reflect visual understanding or linguistic priors inherited in the LLM backbone remains unresolved~\cite{xiao2024can}. This ambiguity motivates Grounded VideoQA, which requires not only a correct answer but also the temporal evidence from which it is derived.

\smallskip
\noindent\textbf{Grounded VideoQA.}
NExT-GQA~\cite{xiao2024can} formalizes visually grounded VideoQA by introducing temporal grounding as an explicit evaluation criterion under weak supervision, where no grounding annotations are available during training. Subsequent methods tackle this challenge from complementary perspectives. TimeCraft~\cite{liu2024timecraft} enforces cycle consistency between causally reversed Q\&A pairs as a self-supervised grounding signal, while CRA-GQA~\cite{chen2025cross} applies causal intervention to disentangle visual and linguistic confounders. QGAC-TR~\cite{xu2024exploring} calibrates temporal attention via contrastive objectives over visual and textual context; on the generative side, Grounded-VideoLLM~\cite{wang2024grounded} augments the vocabulary with discrete temporal tokens, and TOGA~\cite{gupta2025toga} couples timestamp prediction with open-ended answer generation under weak supervision. 
Despite these efforts, our analysis shows that existing methods often produce identical temporal grounding for different questions within the same video, which we attribute to late fusion of independently encoded modalities. GroundFormer addresses this question-invariant grounding by introducing learnable tokens for stronger visuo-lingual alignment under QA weak supervision.

\smallskip
\noindent\textbf{Weakly Supervised Learning for Temporal Grounding.}
In the absence of frame-level annotations, Multiple Instance Learning (MIL)~\cite{dietterich1997solving} treats each video as a bag of temporal instances and identifies relevant segments from video-level supervision alone. 
Prior MIL-based methods~\cite{gao2019wslln,huang2021cross,ma2020vlanet,mithun2019weakly} learn segment-level representations by contrasting matched and mismatched visual-language pairs, achieving competitive performance in video moment retrieval~\cite{mithun2019weakly,ma2020vlanet} and video grounding~\cite{gao2019wslln,huang2021cross}. However, these methods operate on declarative queries (\eg ``A person throws a ball'') that explicitly describe the target event, whereas Grounded VideoQA poses interrogative queries (\eg ``Why did the person throw the ball?'') whose supporting evidence is only implicitly defined by the question--answer relationship. Our approach is the first to adopt a MIL-based formulation for Grounded VideoQA, coupling answer selection with temporal localization through a cross-attention mechanism.
\section{Motivation}
\label{sec:problem_statement}

\noindent\textbf{Observation.}
Most existing Grounded VideoQA methods~\cite{xiao2024can,liu2024timecraft,chen2025cross} share a common grounding structure built upon a Gaussian-based localization module.
Yet, as Fig.~\ref{fig:teaser} illustrates, the previous methods often show \emph{question-invariant grounding}, where models produce nearly identical temporal predictions regardless of the questions.
To quantify this behavior, we introduce two diagnostic metrics.
Pairwise IoU (PIoU) measures the average temporal IoU between predictions for different questions within the same video; high PIoU indicates invariant grounding, while low PIoU indicates question-discriminative grounding.
GT-Pred Correlation (GT-Pred Corr.) complements PIoU by measuring whether predicted overlaps vary in the same direction as ground-truth overlaps, computed as the Pearson correlation between ground-truth and predicted pairwise IoU across the dataset.

We evaluate PIoU and GT-Pred Corr. on NG+~\cite{xiao2024can} and CRA-GQA~\cite{chen2025cross} in~\cref{tab:piou}.
Both methods record PIoU of 85.7 and 88.9, far above the ground-truth PIoU of 21.3, while their GT-Pred Corr. remains near zero (0.005 and 0.006).
These results indicate that their grounding predictions are nearly indistinguishable across questions, motivating our analysis of the common grounding structure underlying this failure.

\begin{table*}[t!]
\caption{Question-invariant grounding analysis on NExT-GQA. Lower PIoU indicates less collapse across questions; higher GT--Pred Corr.\ indicates better agreement with the ground-truth pairwise overlap structure.}
\label{tab:piou}
\centering
\setlength{\tabcolsep}{7pt}
\renewcommand{\arraystretch}{1.15}
\resizebox{0.85\textwidth}{!}{
\begin{tabular}{lcccc}
\toprule
\textbf{Metric} & NG+~\cite{xiao2024can} & CRA-GQA~\cite{chen2025cross} & \textbf{GroundFormer} & Ground Truth \\
\midrule
PIoU $\downarrow$          & 85.7  & 88.9  & \textbf{28.2}  & 21.3 \\
GT--Pred Corr.\ $\uparrow$ & 0.005 & 0.006 & \textbf{0.121} & --   \\
\bottomrule
\end{tabular}
}
\end{table*}

\smallskip
\noindent\textbf{Preliminary.}
Given an untrimmed video $v$, a question $q$, and answer candidates
$A=\{a_1,\ldots,a_{A_n}\}$,
a vision encoder extracts frame-level features
$x_v \in \mathbb{R}^{B\times T\times D_V}$ from $T$ sampled frames, where $B$ is the batch size and $D_V$ the visual feature dimension.
A language encoder produces embeddings
$x_l \in \mathbb{R}^{B\times L_n\times L_l\times D_L}$,
where $L_n$ denotes the number of candidates (question or answer) and $L_l$ is the maximum token length, and $D_L$ the language feature dimension.
The pooled question and answer embeddings, denoted $x_q$ and $x_a$, respectively, are projected to a shared dimension $D$.
The methods then solve:
\begin{equation}    
    a^*, t^* = \operatorname*{arg\,max}_{a \in A}
    \ \Psi\!\left(a \mid v_t, q, A\right)\ \Phi\!\left(t \mid v, q\right),
    \label{eq:dual-style}
\end{equation}
where $\Phi$ is a \emph{grounding module} that estimates a temporal grounding distribution $t$ given $(v,q)$, and $\Psi$ predicts the answer from the localized evidence $v_t$.

A representative grounding module $\Phi(t\mid v,q)$ injects the question into the visual stream as a \emph{single} token.
Let $x_{q_n}$ denote the question token; it is concatenated with $T$ visual tokens to form
$[x_v;\ x_{q_n}] \in \mathbb{R}^{B\times (T+1)\times D}$,
which is then sum-pooled to $\mathbb{R}^{B\times D}$ and passed through $\Phi$ to regress Gaussian parameters for temporal grounding:
\begin{equation}
\begin{gathered}
    (\mu, \sigma^2) = \Phi\!\left(\mathtt{pool}\!\left([x_v;\ x_{q_n}]\right)\right), \qquad
    \mathbf{g}_{\text{gauss}} = \mathcal{N}(\mu, \sigma^2),\\
    \mathbf{g}_{\text{attn}} = \mathtt{Linear}\!\left(\mathbf{g}_{\text{gauss}} \odot x_v\right), \qquad
    t^* = \left(\mathbf{g}_{\text{attn}} \cap \mathbf{g}_{\text{gauss}}\right).
    \label{eq:grounding_module}
\end{gathered}
\end{equation}

\noindent\textbf{Discussion.}
Based on our formulation of the common grounding structure in~\cref{eq:grounding_module}, we identify two structural limitations that make the grounding module $\Phi$ prone to question-invariant behavior.
\begin{enumerate}
    \item \textbf{Modality isolation before $\Phi$.}
    Visual and linguistic branches encode independently, and cross-modal interaction occurs only at a late stage.
    Consequently, frame features carry only limited information about \emph{what the question asks}, leaving the grounding module $\Phi$ to operate on question-agnostic video features.
    \item \textbf{Insufficient linguistic signal inside $\Phi$.}
    The question is injected as a single token among $T$ visual tokens in $[x_v;\ x_{q_n}] \in \mathbb{R}^{B\times (T+1)\times D}$ and is immediately collapsed by pooling.
    With a $1{:}T$ imbalance, the question signal is easily absorbed by the dominant visual stream, making the input to $\Phi$ weakly dependent on the question and falling back to question-invariant grounding.
\end{enumerate}

\begin{figure*}[t]
\centering
\includegraphics[trim=0cm 0cm 0cm 0cm,clip,width=1.0\textwidth]{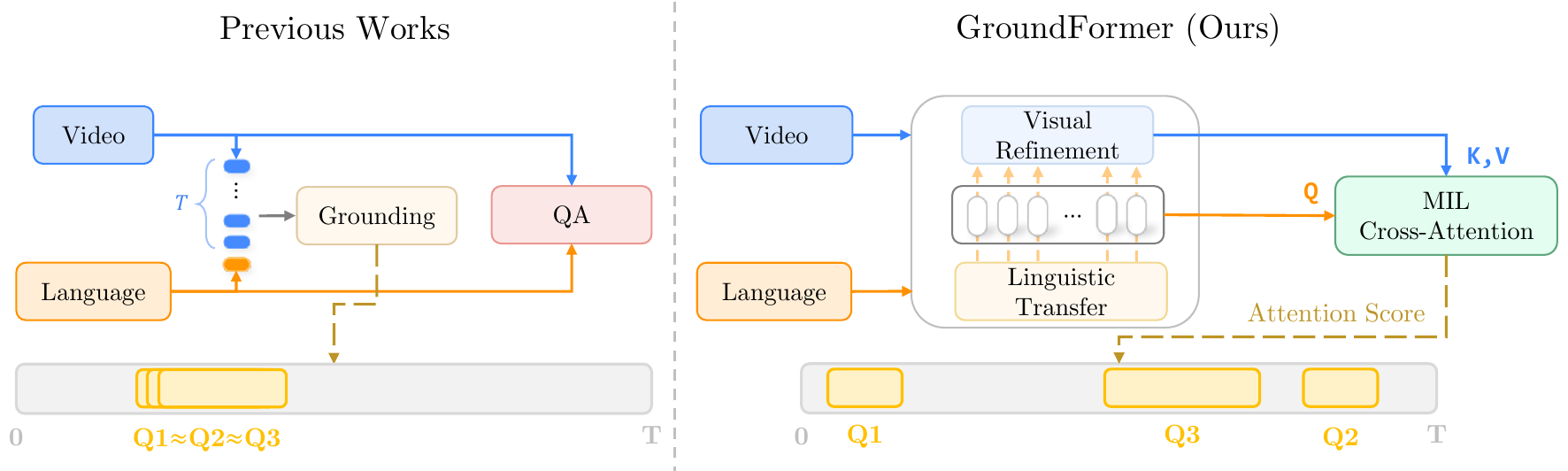}
\caption{\textbf{Overview of GroundFormer.}
\textbf{Left:} Prior methods~\cite{xiao2024can,chen2025cross,liu2024timecraft} weakly inject question information into the grounding module, often collapsing different questions to the same salient interval.
\textbf{Right:} GroundFormer conditions video features on question semantics via communication tokens (\emph{Linguistic Transfer} $\rightarrow$ \emph{Visual Refinement}) and couples answer selection with temporal grounding using MIL Cross-Attention (token queries, video keys/values), producing question-discriminative segments.}
\label{fig:overall}
\end{figure*}

These two limitations compound: modality isolation removes question semantics from the visual features, while the weak linguistic signal prevents $\Phi$ from recovering question dependence later.
As quantified in~\cref{tab:piou}, this drives existing methods toward question-invariant grounding.
GroundFormer breaks this pattern by conditioning visual representations on question semantics before $\Phi$, thereby restoring question-discriminative grounding (\cref{fig:overall}).

\section{Approach}
\label{sec:approach}
We propose GroundFormer, a framework that conditions visual encoding on question semantics prior to the grounding module.
The framework operates as a two-pass pipeline that separates question understanding from answer prediction (\cref{sec:2stage}).
Within each pass, learnable communication tokens bridge the two modalities via Linguistic Transfer and Visual Refinement, and MIL cross-attention derives a question-conditioned grounding signal from answer supervision alone; Gaussian smoothing then refines sparse attention peaks into temporally coherent segments (\cref{sec:groundformer}).
A two-pass pipeline further encourages question-discriminative visual representations during training (\cref{sec:hmcl}), and the model is optimized with classification and hierarchical multi-modal contrastive objectives (\cref{sec:loss}).

\subsection{Two-Pass Pipeline}
\label{sec:2stage}
GroundFormer performs two passes over shared weights. The question pass takes question candidates as input, constructed by pairing the ground-truth question with questions sampled from other videos in the mini-batch as negatives, and identifies the most video-relevant question, encoding visual discriminativity with respect to question semantics. The second pass takes answer candidates as input and produces the answer prediction $\hat{a}$ along with the grounding signal. This disentanglement allows the hierarchical multi-modal contrastive loss to encourage question-discriminative visual representations. Note that despite the two-pass pipeline, the design introduces no additional computational cost at inference, as the two passes share weights and only the answer pass executes at test time.

\subsection{Architecture}
\label{sec:groundformer}
\begin{figure*}[t!]
\centering
\includegraphics[trim=0cm 0cm 0cm 0cm,clip,width=1.0\textwidth]{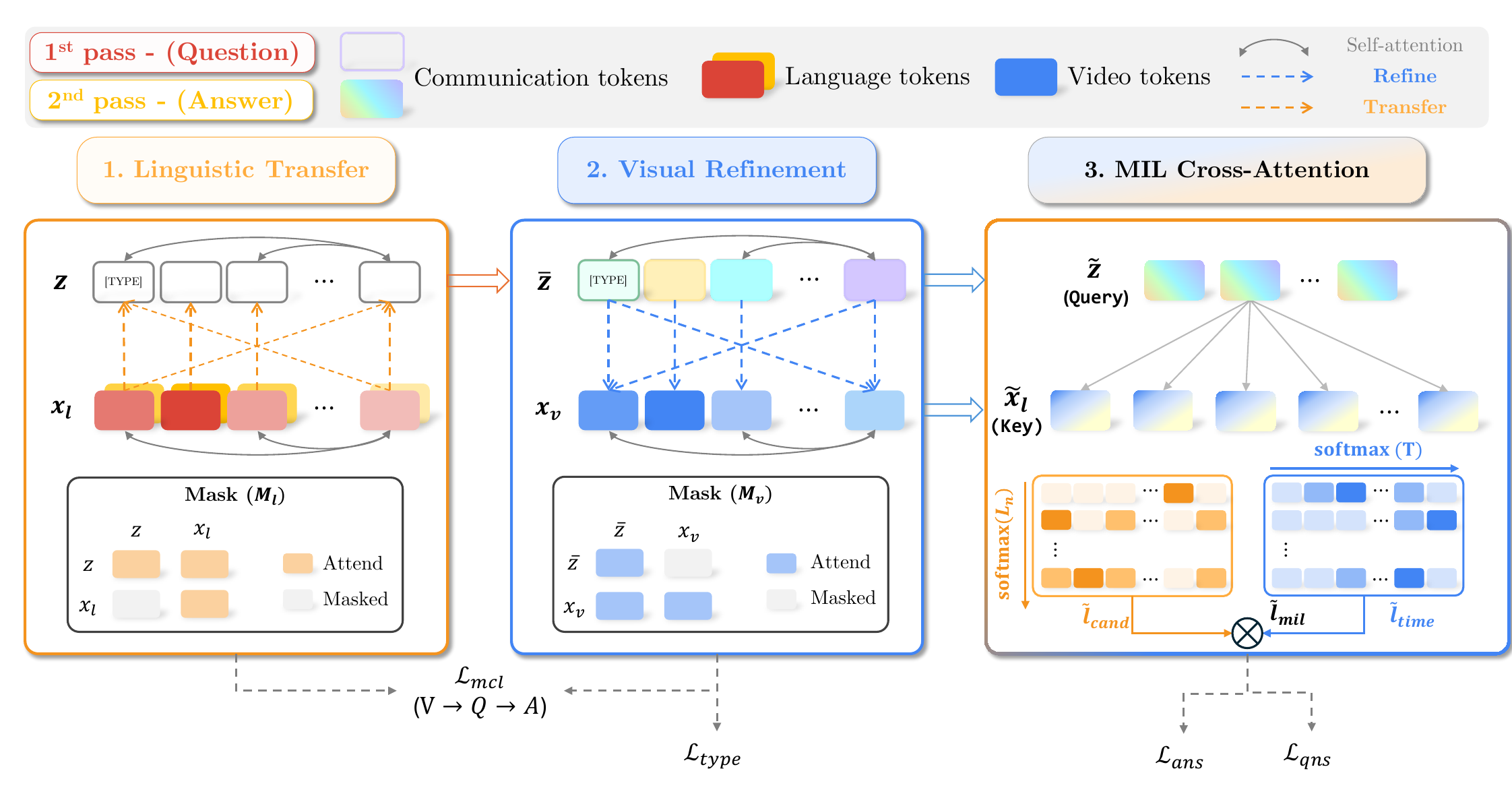}
\caption{\textbf{Details of GroundFormer.}
(1) \textbf{Linguistic Transfer:} communication tokens $z$ attend to language tokens $x_l$ under an asymmetric mask $M_l$ to absorb candidate semantics.
(2) \textbf{Visual Refinement:} video tokens $x_v$ attend to the distilled tokens $\bar{z}$ under $M_v$, producing question-conditioned features $\tilde{x}_v$ while keeping tokens uncontaminated.
(3) \textbf{MIL Cross-Attention:} token queries attend to video keys/values to obtain attention logits, which are factorized into candidate and temporal distributions ($\tilde{\ell}_{cand}$, $\tilde{\ell}_{time}$) whose product yields MIL scores and an implicit grounding signal.
The model is trained with losses for question matching (1st pass) and answer prediction (2nd pass), together with type and multi-modal contrastive objectives.}
\label{fig:details}
\end{figure*}

GroundFormer consists of two core components: learnable communication tokens that transfer question semantics into visual representations, and MIL cross-attention that derives a question-conditioned grounding signal from answer supervision. \cref{fig:details} illustrates the overall architecture and attention flow.

\smallskip
\noindent\textbf{Communication Tokens.}
GroundFormer introduces a set of communication tokens
$z=[z_{\mathtt{type}};\ z_{\mathtt{query}}]\in\mathbb{R}^{B\times(1+N)\times D}$
as a bottleneck between two modalities.
The $\mathtt{[TYPE]}$ token $z_{\mathtt{type}}\in\mathbb{R}^{B\times1\times D}$ encodes a coarse question category and is supervised by an auxiliary loss (\cref{sec:loss});
this coarse prior guides the remaining tokens to specialize across reasoning patterns.
The $N$ query tokens $z_{\mathtt{query}}\in\mathbb{R}^{B\times N\times D}$ capture fine-grained question semantics and propagate the captured intent to the visual representation through two steps: \emph{Linguistic Transfer} and \emph{Visual Refinement}.

\smallskip
\textit{Linguistic Transfer} first transfers candidate semantics into the communication tokens via asymmetric masked self-attention. 
A transformer layer concatenates $z$ with $x_l$ along the token dimension, allowing $z$ to attend to both $z$ and $x_l$ while $x_l$ attends only to itself:
\begin{equation}
    \bar{z} = \texttt{SA}_1([z;\ x_l], M_l)\big|_{z}.
    \label{eq:intent}
\end{equation}
The asymmetric mask ensures that $\bar{z}$ captures candidate semantics while preventing the communication tokens from corrupting the language representations.

\smallskip
\textit{Visual Refinement} propagates the captured intent $\bar{z}$ into the visual stream via a second asymmetric masked self-attention layer.
Concatenating $x_v$ with $\bar{z}$, the layer allows $x_v$ to attend to both $x_v$ and $\bar{z}$ while $\bar{z}$ attends only to itself
\begin{equation}
    \tilde{x}_v = \texttt{SA}_2([x_v;\ \bar{z}], M_v)\big|_{x_v}, \qquad
    \tilde{z}   = \texttt{SA}_2([x_v;\ \bar{z}], M_v)\big|_{z}.
    \label{eq:refine}
\end{equation}
Each frame in $\tilde{x}_v$ now encodes candidate-specific information, resolving the modality isolation. The asymmetric mask keeps $\tilde{z}$ independent of visual content, and the updated $\tilde{z}_{\texttt{query}}$ serves as queries for the subsequent grounding step.

\smallskip
\noindent\textbf{Weakly-Supervised Grounding via MIL Cross-Attention.}
We generate a grounding signal from QA supervision alone using MIL cross-attention mechanism.
We stack candidates along the $L_n$ dimension, yielding intent queries
$\tilde{z}_{\mathtt{query}}\in\mathbb{R}^{B\times L_n\times N\times D}$
and question-conditioned visual tokens
$\tilde{x}_v\in\mathbb{R}^{B\times L_n\times T\times D}$.
We compute cross-attention logits:
\begin{equation}
\begin{gathered}
    \tilde{\ell} = \frac{(\tilde{z}_{\mathtt{query}}W_Q) \cdot (\tilde{x}_vW_K)^{\intercal}}{\sqrt{D_k}}
    \in \mathbb{R}^{B\times L_n\times N\times T},\\
    \tilde{\ell}_{time} = \mathtt{softmax}_{T}(\tilde{\ell}), \qquad
    \tilde{\ell}_{cand} = \mathtt{softmax}_{L_n}(\tilde{\ell}),\\
    \tilde{\ell}_{mil} = \sum_{t=1}^{T}\left(\tilde{\ell}_{time}\odot \tilde{\ell}_{cand}\right)
    \in \mathbb{R}^{B\times L_n\times N}.
    \label{eq:mil}
\end{gathered}
\end{equation}
Here, $\tilde{\ell}_{time}$ distributes probability over frames for each candidate, acting as the temporal evidence signal, while $\tilde{\ell}_{cand}$ distributes probability over candidates at each frame.
Their element-wise product assigns a high score to a candidate only when it is supported by frames that also receive high temporal attention; averaging over time yields the MIL logit $\tilde{\ell}_{mil}$.
This factorization binds the candidate score to its supporting frames, so $\tilde{\ell}_{time}$ serves as a grounding signal derived from answer supervision.
The cross-attention output aggregates visual evidence weighted by $\tilde{\ell}_{time}$, and average pooling over the $N$ query tokens produces $\hat{z} \in \mathbb{R}^{B \times L_n \times D}$, which is fed to the classifier heads.

\smallskip
\noindent\textbf{Grounding Refinement via Gaussian Kernel.}
\label{sec:gkernel}
MIL Cross-Attention produces a candidate-conditioned temporal distribution $\tilde{\ell}_{time}$.
In the answer pass, we take the temporal attention associated with the predicted answer $\hat{a}$ as the raw grounding signal $\mathbf{g}_{attn} \in \mathbb{R}^{B \times T}$.
MIL factorization often concentrates attention on the most discriminative frames, resulting in sharply peaked signals that can fragment the predicted temporal segment. We therefore apply a 1-D Gaussian kernel to smooth $\mathbf{g}_{attn}$ into a temporally continuous segment.
\begin{equation}
    \mathcal{G}_{\sigma}(k) = \exp\!\left(-\frac{k^2}{2\sigma^2}\right), \qquad
    \mathbf{g}_{gauss} = \mathcal{G}_{\sigma} * \mathbf{g}_{attn},
    \label{eq:gauss_grounding}
\end{equation}
where $k$ indexes the kernel position, $*$ denotes 1-D convolution over the $T$ frames, and $\sigma$ controls the smoothing extent.
We use $\mathbf{g}_{gauss}$ to derive the final grounded segment.

\subsection{Hierarchical Multi-modal Contrastive Learning}
\label{sec:hmcl}
Unlike declarative queries that directly describe a target event, interrogative queries in Grounded VideoQA require an intermediate reasoning step: the model must first identify the relevant visual evidence before arriving at the answer. This two-step reasoning naturally motivates a hierarchical alignment structure across the two-pass pipeline. In the first pass, video embeddings are aligned with question embeddings (V$\rightarrow$Q), grounding visual representations in question semantics. In the second pass, question embeddings are aligned with answer embeddings (Q$\rightarrow$A), binding the answer to the question-conditioned visual evidence. Together, V$\rightarrow$Q$\rightarrow$A alignment pulls video, question, and answer embeddings into a coherent embedding space progressively, using in-batch candidates as negative samples.
\subsection{Training Objectives}
\label{sec:loss}
GroundFormer is trained end-to-end with candidate-level supervision only. Each pass is supervised by standard cross-entropy over its candidates using the pooled representation $\hat{z}$. We additionally apply cross-entropy on the MIL logits $\tilde{\ell}_{mil}$ to ensure that the factorized scoring aligns with the ground-truth candidate:
\begin{equation}
\begin{gathered}
    \mathcal{L}_{qns} =
    \mathtt{CE}\!\left(f_Q(\hat{z}_Q), y_q\right)
    + \mathtt{CE}\!\left(\tilde{\ell}_{mil}^{Q}, y_q\right),\\
    \mathcal{L}_{ans} =
    \mathtt{CE}\!\left(f_A(\hat{z}_A), y_a\right)
    + \mathtt{CE}\!\left(\tilde{\ell}_{mil}^{A}, y_a\right),
    \label{eq:ce_loss}
\end{gathered}
\end{equation}
where $f_Q$ and $f_A$ are the question and answer classifier heads, $y_q$ and $y_a$ are the ground-truth indices.

The hierarchical multi-modal contrastive loss aligns embeddings progressively across the two passes. Let $\tilde{x}_v^i\in\mathbb{R}^{D}$ denote the pooled video embedding from the question pass for the $i$-th sample in a mini-batch of size $B$, $\bar{x}_{q}^{+,i}\in\mathbb{R}^{D}$ its ground-truth question embedding, and $\tau$ a temperature parameter. All $B\times Q_n$ question embeddings in the mini-batch form the contrastive pool, where $\bar{x}_q^{j,m}$ is the $m$-th question-candidate embedding of the $j$-th sample. So the (V$\rightarrow$Q) alignment is:
\begin{equation}
  \mathcal{L}_{vq} = -\frac{1}{B}\sum_{i=1}^{B}
    \log \frac{
      \exp\bigl(\mathrm{sim}(\tilde{x}_v^i,\, \bar{x}_{q}^{+,i}) \,/\, \tau\bigr)
    }{
      \sum_{j=1}^{B}\sum_{m=1}^{Q_n}
      \exp\bigl(\mathrm{sim}(\tilde{x}_v^i,\, \bar{x}_{q}^{j,m}) \,/\, \tau\bigr)
    }.
  \label{eq:loss_vq}
\end{equation}
Using the same anchor $\bar{x}_{q}^{+,i}$, the (Q$\rightarrow$A) alignment is defined against all $B\times A_n$ answer embeddings:
\begin{align}
  \mathcal{L}_{qa} &= -\frac{1}{B}\sum_{i=1}^{B}
    \log \frac{
      \exp\bigl(\mathrm{sim}(\bar{x}_{q}^{+,i},\, \bar{x}_{a}^{+,i}) \,/\, \tau\bigr)
    }{
      \sum_{j=1}^{B}\sum_{m=1}^{A_n}
      \exp\bigl(\mathrm{sim}(\bar{x}_{q}^{+,i},\, \bar{x}_{a}^{j,m}) \,/\, \tau\bigr)
    }
  \label{eq:loss_qa}\\
  \mathcal{L}_{mcl} &= \mathcal{L}_{vq} + \mathcal{L}_{qa}
  \label{eq:loss_mcl}
\end{align}

The $\mathtt{[TYPE]}$ is supervised with an auxiliary classification loss:
\begin{equation}
    \mathcal{L}_{type} = \mathtt{CE}\!\left(f_T(\tilde{z}_{type}), y_{type}\right),
    \label{eq:type_loss}
\end{equation}
where $f_T$ is a linear classifier and $y_{type}$ is the question category label provided by each benchmark~\cite{xiao2021next,wu2021star_situated_reasoning}. 

The total loss combines all objectives:
\begin{equation}
    \mathcal{L} = \mathcal{L}_{ans} + \mathcal{L}_{qns} + \mathcal{L}_{mcl} + \mathcal{L}_{type}.
    \label{eq:total_loss}
\end{equation}
\section{Experiment}
\label{sec:exp}

\subsection{Experimental Setup}
\noindent\textbf{Datasets.}
We evaluate GroundFormer on two popular grounded VideoQA benchmarks. NExT-GQA~\cite{xiao2024can} targets causal and temporal reasoning through human-annotated questions, while STAR~\cite{wu2021star_situated_reasoning} targets situated reasoning through programmatically generated questions.
\begin{itemize}[leftmargin=1.2em,itemsep=0.2ex,topsep=0.2ex,parsep=0pt,partopsep=0pt]
    \item \textbf{NExT-GQA}~\cite{xiao2024can} extends NExT-QA~\cite{xiao2021next} with temporal grounding annotations for the validation and test splits (10.5K annotated QA pairs), enabling evaluation of both answer correctness and evidence localization.
    Compared to short-clip VideoQA benchmarks~\cite{choi2021dramaqa,jang2017tgif,min2024morevqa} ($\sim$10\,s), it contains longer untrimmed videos (avg.\ $\sim$44\,s) and emphasizes causal and temporal reasoning under weak supervision, spanning five question types: Causal-Why, Causal-How, Temporal-Before\&After, Temporal-When, and Temporal-Present.

    \item \textbf{STAR}~\cite{wu2021star_situated_reasoning} is a situated reasoning benchmark with 4,901 videos and 60,206 QA pairs, each annotated with a temporal segment.
    Videos average $\sim$30\,s with a segment-to-video ratio of $\sim$0.4, roughly twice that of NExT-GQA. Questions are categorized into four types: Interaction, Sequence, Prediction, and Feasibility.
\end{itemize}

\noindent\textbf{Evaluation Metrics.}
We report metrics for both answering and grounding. For QA, we report Acc@VQA, the percentage of correctly answered questions. For localization, we report mIoU and mIoP between the predicted and ground-truth temporal segments. Since Grounded VideoQA requires both a correct answer \textit{and} faithful grounding, we use Acc@GQA as the primary metric. Following NExT-GQA, Acc@GQA counts a prediction as correct only if the answer is correct and the grounded segment achieves IoP $\ge 0.5$.

\smallskip
\noindent\textbf{Implementation Details.}
We implement GroundFormer in PyTorch.
For video representation, we uniformly sample $T{=}32$ frames per video and extract features using a frozen CLIP ViT-L/14 vision encoder~\cite{radford2021learning}.
For text, we use a frozen RoBERTa-base encoder~\cite{liu2019roberta} to embed questions and answer candidates.
The GroundFormer block consists of 4 transformer layers. We set the number of learnable query tokens to $N{=}32$ and the Gaussian kernel to $\sigma{=}2$. We train for 30 epochs with batch size 16 using AdamW~\cite{loshchilov2017decoupled}, an initial learning rate of $10^{-5}$, and a cosine decay schedule.
All experiments are conducted on RTX 3090 GPUs.

\smallskip
\noindent\textbf{Baselines.}
We compare our method against grounding methods: PH~\cite{xiao2024can}, NG+~\cite{xiao2024can}, TimeCraft~\cite{liu2024timecraft}, and CRA-GQA~\cite{chen2025cross}, while varying their base architectures with different capacities: Temp[CLIP]~\cite{xiao2024can} (130--145M) and FrozenBiLM~\cite{yang2022zero} (1.2B). We also compare our method with QGAC-TR~\cite{xu2024exploring} using its own architecture for Grounded VideoQA.

\subsection{Comparison With State-of-the-Art Methods}
We demonstrate the robustness of GroundFormer on two Grounded VideoQA benchmarks~\cite{xiao2024can, wu2021star_situated_reasoning}. Our framework achieves new state-of-the-art Acc@GQA and temporal grounding across both datasets regardless of network capacity.

\smallskip
\noindent\textbf{NExT-GQA.}
\cref{tab:sota_nextgqa} compares GroundFormer with prior methods on the NExT-GQA test set.
GroundFormer achieves the best Acc@GQA of 21.5, setting a new state of the art with consistent gains in localization quality across all grounding metrics and thresholds.
While FrozenBiLM variants attain higher Acc@VQA (up to 74.7) due to their larger capacity, GroundFormer yields substantially stronger grounded performance: it improves Acc@GQA (21.5 vs.\ 18.8), mIoP (34.0 vs.\ 26.5), and mIoU (17.5 vs.\ 13.5) over the best FrozenBiLM baseline, despite using only 17.6\% of its parameters.

\begin{table*}[t]
\caption{
Comparison with state-of-the-art methods on NExT-GQA~\cite{xiao2024can}.
}
\label{tab:sota_nextgqa}
\centering
\setlength{\tabcolsep}{4.5pt}
\renewcommand{\arraystretch}{1.05}
\resizebox{\textwidth}{!}{
\begin{tabular}{c|l|c|c|c|ccc|ccc}
\toprule
Arch. & \multicolumn{1}{c|}{Method} & Param. & \textbf{Acc@GQA} & Acc@VQA & mIoP & TIoP@0.3 & TIoP@0.5 & mIoU & TIoU@0.3 & TIoU@0.5 \\
\midrule\midrule
\multirow{4}{*}{Temp[CLIP]~\cite{xiao2024can}}
& PH~\cite{xiao2024can}          &130M  & 15.2 & 59.4 & 25.4 & 28.2 & 25.5 &  6.6 &  9.3 &  4.1 \\
& NG+~\cite{xiao2024can}         &130M  & 16.0 & 60.2 & 25.7 & 31.4 & 25.5 & 12.1 & 17.5 &  8.9 \\
& TimeCraft~\cite{liu2024timecraft} &130M & 18.2 & \textbf{65.6} & 28.1 & 35.1 & 27.8 & 15.6 & 21.2 &  9.6 \\
& CRA-GQA~\cite{chen2025cross}   &145M    & 18.2 & 61.1 & 28.6 & 34.3 & 28.5 & 14.2 & 21.4 & 10.6 \\
\midrule
\multirow{4}{*}{FrozenBiLM~\cite{yang2022zero}}
& PH~\cite{xiao2024can}          & 1.2B & 15.8 & 69.1 & 22.7 & 25.8 & 22.1 &  7.1 & 10.0 &  4.4 \\
& NG+~\cite{xiao2024can}         & 1.2B & 17.5 & 70.8 & 24.2 & 28.5 & 23.7 &  9.6 & 13.5 &  6.1 \\
& TimeCraft~\cite{liu2024timecraft} &1.2B & 18.5 & 74.7 & 26.3 & 32.7 & 24.9 & 13.2 & 18.6 &  8.4 \\
& CRA-GQA~\cite{chen2025cross}   &1.2B    & 18.8 & 70.2 & 26.5 & 32.6 & 25.9 & 13.5 & 20.1 &  9.6 \\
\midrule
\multicolumn{2}{c|}{QGAC-TR~\cite{xu2024exploring}} &-- & 18.3 & 63.6 & 28.3 & 32.8 & 27.7 & 15.7 & 18.6 & 11.7 \\
\multicolumn{2}{c|}{\textbf{GroundFormer}} &211M & \textbf{21.5} & 61.7 & \textbf{34.0} & \textbf{41.5} & \textbf{33.7} & \textbf{17.5} & \textbf{26.5} & \textbf{12.8} \\
\bottomrule
\end{tabular}}
\end{table*}

\begin{figure*}[t]
\centering
\includegraphics[trim=0cm 0cm 0cm 0cm,clip,width=1.0\textwidth]{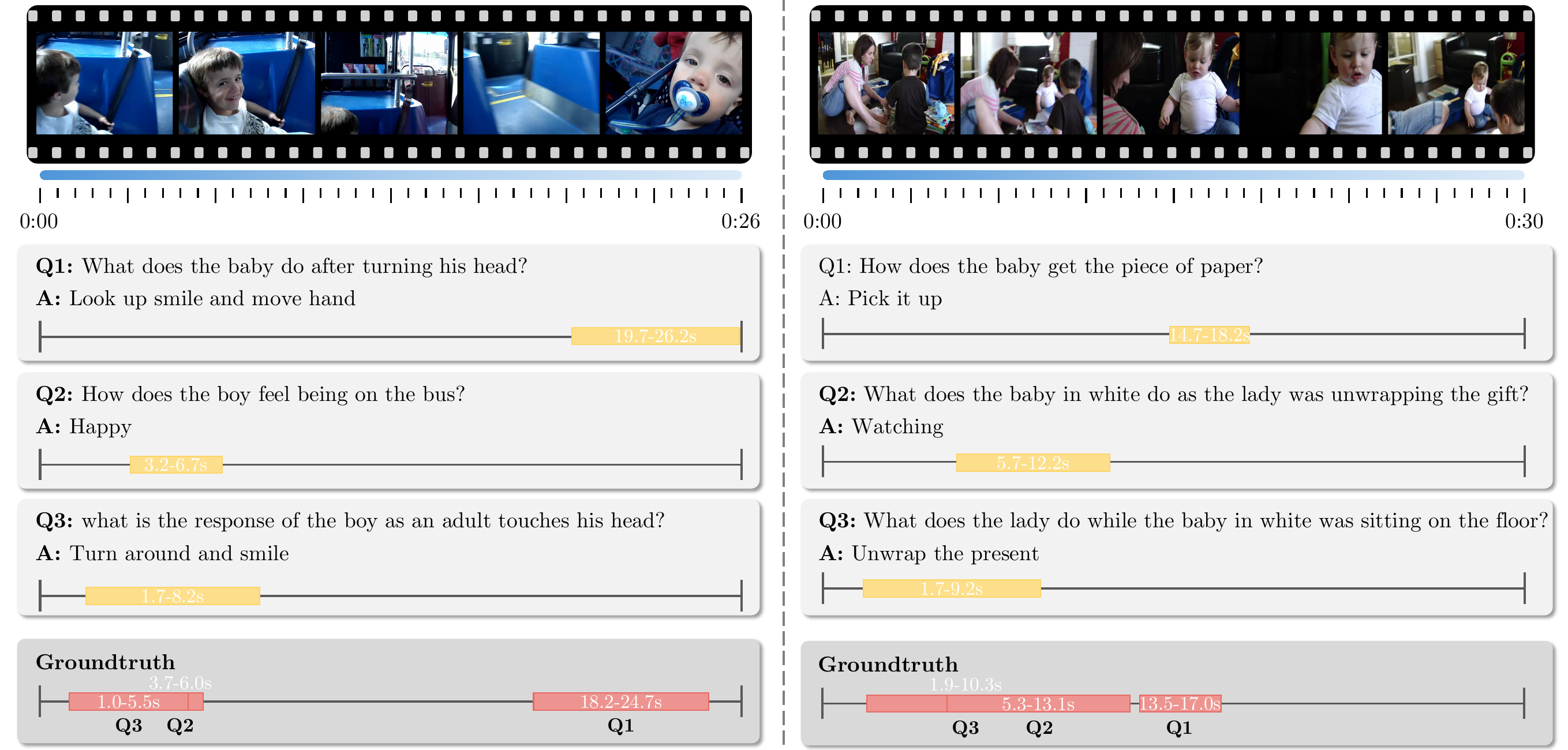}
\caption{\textbf{Qualitative comparison on NExT-GQA.}
For each video, we show three questions (Q1--Q3) with their answers and the predicted grounding intervals, along with the ground-truth segments (bottom).
GroundFormer produces distinct, question-conditioned temporal segments that follow the ground-truth ordering, whereas prior methods collapse different questions onto similar intervals.}
\label{fig:qualitative}
\end{figure*}

\cref{fig:qualitative} further illustrates the benefit of GroundFormer's question-conditioned grounding capability.
In the left example, three questions target temporally distinct events: Q3 asks about a reaction near the beginning, Q2 about a feeling in the middle, and Q1 about an action toward the end.
GroundFormer localizes each question to a different temporal region that aligns with the corresponding ground-truth segment.
In the right example, three questions again refer to separate moments, and GroundFormer produces three non-overlapping predictions that match the ground-truth ordering.
In both cases, NG+ and CRA-GQA collapse all predictions onto a single interval regardless of the question.
Thus, our result confirms that GroundFormer ensures correct answers are derived from accurately localized temporal evidence compared to baseline methods.

\noindent\textbf{STAR.}
\cref{tab:star} demonstrates the generalization capability of GroundFormer on STAR benchmark. Compared to existing methods under both TempCLIP~\cite{xiao2024can} and FrozenBiLM~\cite{yang2022zero} architectures, GroundFormer establishes a new state-of-the-art with an Acc@GQA of 30.7. The improvement is driven by robust temporal grounding, as shown by an mIoP@0.5 of 49.3 and mIoU@0.5 of 15.8.

\begin{table*}[t]
\caption{Comparison with state-of-the-art methods on STAR~\cite{wu2021star_situated_reasoning}.}
\label{tab:star}
\centering
\setlength{\tabcolsep}{4.5pt}
\renewcommand{\arraystretch}{1.05}
\resizebox{0.75\textwidth}{!}{
\begin{tabular}{c|l|c|c|cc}
\toprule
Arch. & \multicolumn{1}{c|}{Method} & \textbf{Acc@GQA} & Acc@VQA & mIoP@0.5 & mIoU@0.5 \\
\midrule\midrule
\multirow{2}{*}{TempCLIP~\cite{xiao2024can}}
& NG+~\cite{xiao2024can}         & 24.4 & 57.3 & 41.4 &  4.7 \\
& CRA-GQA~\cite{chen2025cross}   & 26.8 & 58.6 & 44.5 &  5.5 \\
\midrule
\multirow{2}{*}{FrozenBiLM~\cite{yang2022zero}}
& NG+~\cite{xiao2024can}         & 25.8 & 60.1 & 40.9 &  7.8 \\
& CRA-GQA~\cite{chen2025cross}   & 27.8 & 60.5 & 43.1 &  5.1 \\
\midrule
\multicolumn{2}{c|}{\textbf{GroundFormer}} & \textbf{30.7} & \textbf{61.1} & \textbf{49.3} & \textbf{15.8} \\
\bottomrule
\end{tabular}}
\end{table*}

\begin{table*}[t]
\caption{Ablation on NExT-GQA. We factorize design choices into structural components (two-pass training, communication tokens) and auxiliary components.
}
\label{tab:ablation_all}
\centering
\setlength{\tabcolsep}{3pt}
\renewcommand{\arraystretch}{1.05}
\resizebox{\textwidth}{!}{
\begin{tabular}{cc|cccc|cc|ccc|ccc}
\toprule
\multicolumn{2}{c|}{\textbf{Structure}} &
\multicolumn{4}{c|}{\textbf{Auxiliary Components}} &
\multicolumn{2}{c|}{\textbf{QA}} &
\multicolumn{3}{c|}{\textbf{Localization (IoP)}} &
\multicolumn{3}{c}{\textbf{Localization (IoU)}} \\
\cmidrule(lr){1-2}\cmidrule(lr){3-6}\cmidrule(lr){7-8}\cmidrule(lr){9-11}\cmidrule(lr){12-14}
\makecell{Two-pass\\train} &
\makecell{Comm.\\tokens} &
$\mathcal{L}_{type}$ & $\mathcal{L}_{mcl}$ & $\mathcal{G}_{\sigma}$ & MIL &
\textbf{Acc@GQA} & Acc@VQA &
mIoP & TIoP@0.3 & TIoP@0.5 &
mIoU & TIoU@0.3 & TIoU@0.5 \\
\midrule\midrule
\cmark &  &  &  &  &  &
14.8 & 58.5 & 26.5 & 33.3 & 24.8 & 14.7 & 21.9 & 10.6 \\
\cmark & \cmark &  &  &  &  &
18.6 & 59.8 & 29.6 & 37.0 & 28.6 & 17.0 & 25.8 & 13.0 \\
\cmark & \cmark & \cmark &  &  &  &
19.0 & 60.4 & 30.5 & 37.9 & 30.2 & 16.9 & 25.8 & 13.2 \\
\cmark & \cmark & \cmark & \cmark &  &  &
20.1 & 61.2 & 32.0 & 39.5 & 30.3 & 17.2 & 25.8 & 12.5 \\
\cmark & \cmark & \cmark & \cmark & \cmark &  &
20.8 & 61.2 & 32.4 & 39.3 & 32.0 & 16.5 & 25.0 & 12.0 \\
\midrule
& \cmark &  &  &\cmark  &\cmark  &
15.7 & 58.2 & 26.7 & 32.8 & 25.9 & 14.6 & 21.8 & 10.8 \\
\cmark & \cmark & \cmark & \cmark & \cmark & \cmark &
\textbf{21.5} & \textbf{61.7} & \textbf{34.0} & \textbf{41.5} & \textbf{33.7} &
\textbf{17.5} & \textbf{26.5} & \textbf{12.8} \\
\bottomrule
\end{tabular}}
\end{table*}

\begin{table*}[t]
\caption{Acc@GQA (\%) by question type on NExT-GQA.}
\label{tab:qtype}
\centering
\setlength{\tabcolsep}{7pt}
\renewcommand{\arraystretch}{1.05}
\resizebox{0.8\textwidth}{!}{
\begin{tabular}{l cc ccc}
\toprule
\multirow{2}{*}{Method} & \multicolumn{2}{c}{Causal} & \multicolumn{3}{c}{Temporal} \\
\cmidrule(lr){2-3} \cmidrule(lr){4-6}
& Why & How & Before \& After & When & Present \\
\midrule\midrule
NG+~\cite{xiao2024can}       & 16.9 & 17.2 & 12.0 & 17.5 & 10.8 \\
\textbf{GroundFormer (Ours)} & \textbf{22.2} & \textbf{23.1} & \textbf{17.3} & \textbf{25.8} & \textbf{12.9} \\
\bottomrule
\end{tabular}}
\end{table*}

\begin{figure*}[t]
\centering
\includegraphics[trim=0cm 0cm 0cm 0cm,clip,width=1.0\textwidth]{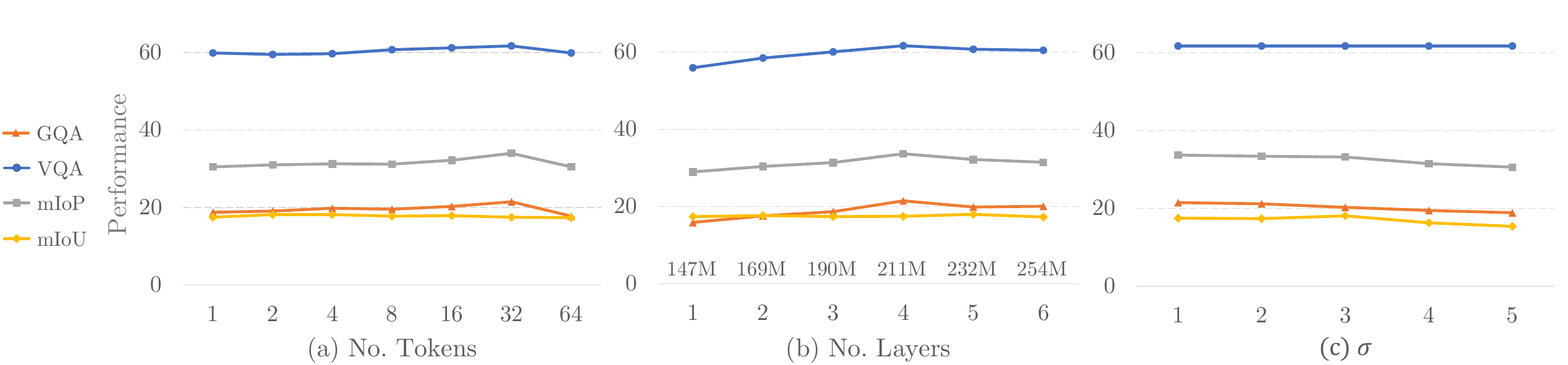}
\caption{\textbf{Hyperparameter sensitivity on NExT-GQA.}
(a) Number of learnable query tokens $N$,
(b) number of Transformer layers in the GroundFormer block,
and (c) Gaussian smoothing width $\sigma$.
We report Acc@VQA, Acc@GQA, mIoP, and mIoU.}
\label{fig:hyperparameter}
\end{figure*}

\subsection{Analysis}
\noindent\textbf{How does each module contribute?}
\cref{tab:ablation_all} summarizes how each design choice contributes to question-conditioned grounding on NExT-GQA.
Starting from a baseline that uses a standard cross-attention grounding head (no communication tokens or auxiliary objectives), we obtain 14.8 Acc@GQA and 26.5 mIoP.
Introducing communication tokens with \emph{Linguistic Transfer} and \emph{Visual Refinement} yields a large jump to 18.6 Acc@GQA and 29.6 mIoP, confirming that conditioning video features on question semantics \emph{before} localization is critical.
Adding our auxiliary objectives provides further gains: question-type supervision and the hierarchical contrastive loss improve Acc@GQA to 19.0 and 20.1, respectively, indicating that coarse reasoning priors and cross-modal alignment are complementary.
Finally, adding the localization heads (Gaussian smoothing and MIL) strengthens localization and leads to the best overall performance.

We also ablate the \emph{two-pass training} strategy in \cref{tab:ablation_all}.
The second-to-last row reports a \emph{single-pass} variant trained using only the answer pass; in this setting, the question-pass objectives (\eg $\mathcal{L}_{qns}$, $\mathcal{L}_{type}$ and the V$\rightarrow$Q term in $\mathcal{L}_{mcl}$) are not applicable.
This single-pass training results in 15.7 Acc@GQA and 26.7 mIoP.
In contrast, enabling two-pass training by adding the question pass activates the full hierarchical objectives and improves performance to 21.5 Acc@GQA and 34.0 mIoP (last row).
Importantly, this gain comes without inference overhead, since only the answer pass is executed at test time.

\smallskip
\noindent\textbf{How does question type shape grounding?}
\cref{tab:qtype} reports Acc@GQA performance separately for each question types defined in NExT-QA~\cite{xiao2021next}: Causal (Why, How) and Temporal (Before \& After, When, Present). GroundFormer outperforms NG+ across all five types, confirming that our question conditioning improves grounding across question types. In addition, we found that the largest gains appear on Temporal-When (+8.3) and Causal-How (+5.9), both of which demand precise temporal discrimination to locate a specific moment. 

\smallskip
\noindent\textbf{Hyperparameter sensitivity.}
\cref{fig:hyperparameter} reports performance under different numbers of learnable tokens $N$, Transformer layers, and Gaussian kernel width $\sigma$.
The Acc@GQA and mIoP peak at $N$ = 32, after which additional tokens yield diminishing returns while increasing parameter count. Performance increases steadily up to 4 layers and saturates thereafter. For $\sigma$, the optimum lies at $\sigma$=2; larger values over-smooth the temporal signal and degrade mIoU. We adopt $N$ = 32, 4 layers, and $\sigma$ = 2 throughout all other experiments.

\begin{figure*}[t]
\centering
\includegraphics[trim=0cm 0cm 0cm 0cm,clip,width=1.0\textwidth]{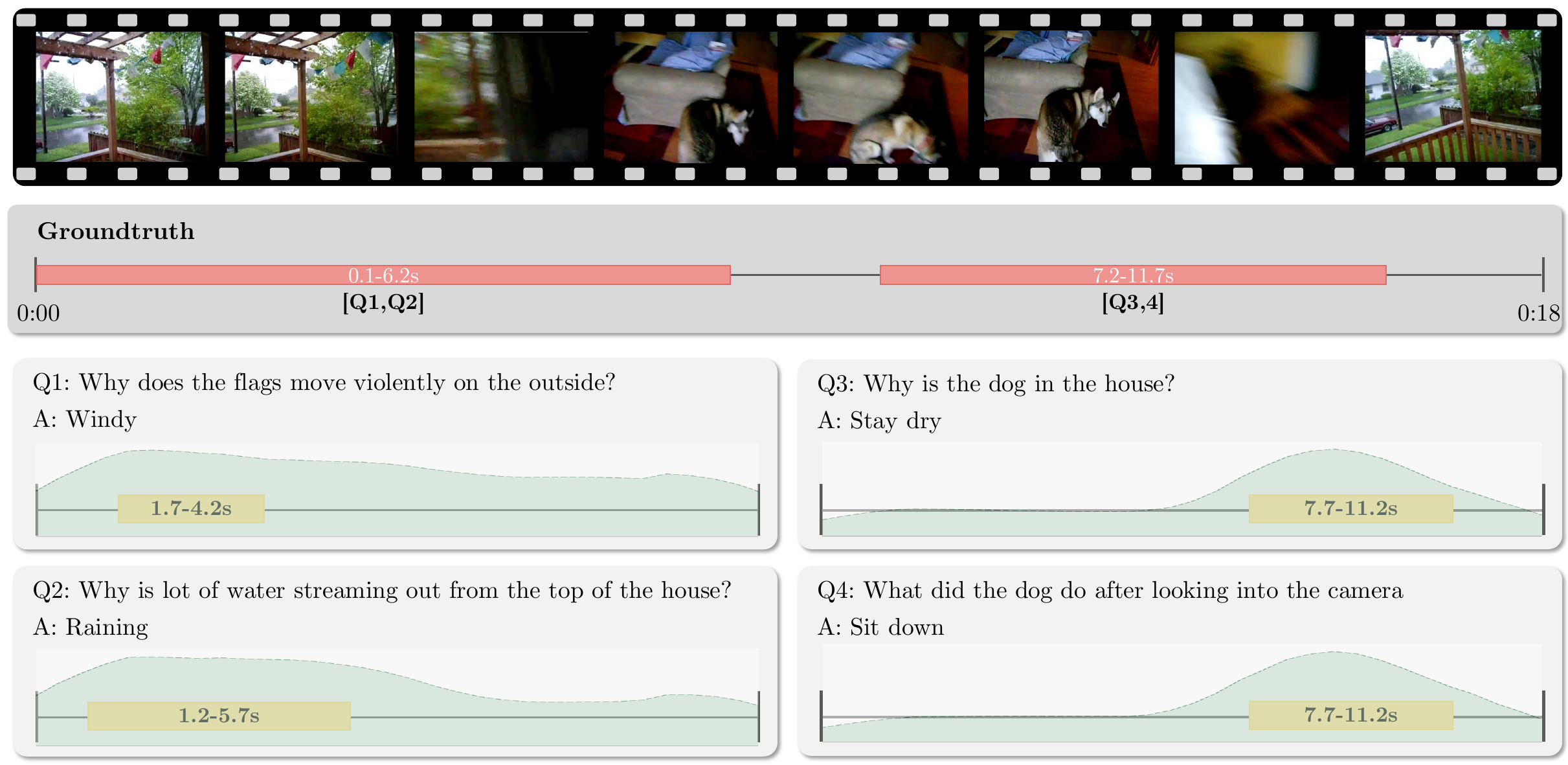}
\caption{\textbf{Grounding for semantically similar questions.}
In the same video, GroundFormer groups Q1--Q2 (outdoor weather) into an early interval and Q3--Q4 (indoor dog event) into a later interval, matching the ground-truth split. Shaded spans indicate predicted grounded segments.}
\label{fig:similar}
\end{figure*}

\smallskip
\noindent\textbf{How does grounding behave for similar questions?}
\cref{fig:similar} shows that GroundFormer produces \emph{consistent} temporal grounding for semantically related questions while remaining \emph{discriminative} across different events within the same video.
Q1 (windy flags) and Q2 (rain streaming from the roof) both refer to the outdoor weather, and GroundFormer grounds them in the early portion of the video (0.1--6.2\,s), consistent with the ground-truth intervals (Q1: 1.7--4.2\,s; Q2: 1.2--5.7\,s).
In contrast, Q3 (dog stays indoors to stay dry) and Q4 (dog sits after looking into the camera) target the later indoor event, and the predicted grounding shifts to 7.2--11.7\,s, aligning with the shared ground-truth segment (7.7--11.2\,s).
These results suggest that GroundFormer’s temporal grounding follows question semantics rather than collapsing onto a single salient moment.

\section{Conclusion}
\label{sec:conclusion}

We identified \emph{question-invariant grounding} as a key failure mode in weakly-supervised Grounded VideoQA: existing methods produce nearly identical temporal predictions regardless of question intent, a consequence of modality isolation and weak question injection in the grounding module.
To address this, we proposed GroundFormer, which conditions visual tokens on question semantics before localization via asymmetric masked attention and couples answer selection with temporal grounding through MIL Cross-Attention.
To further bind grounding to question intent, we introduce a hierarchical multi-modal contrastive objective that aligns video, question, and answer embeddings progressively along a V$\rightarrow$Q$\rightarrow$A.
As a result, GroundFormer achieves state-of-the-art Acc@GQA on both NExT-GQA and STAR, with PIoU and GT-Pred Correlation analyses confirming substantially more question-discriminative grounding.

\section*{Acknowledgements}
This work was supported by Institute of Information $\&$ communications Technology Planning $\&$ Evaluation (IITP) grant funded by the Korea government (MSIT) (No.RS-2025-25443318, Physically-grounded Intelligence: A Dual Competency Approach to Embodied AGI through Constructing and Reasoning in the Real World) and the NRF grant (No. RS-2023-00208506).
Junhyug Noh was supported by the National Research Foundation of Korea (NRF) grant funded by the Korean government (MSIT) (RS-2026-25499022), and Global -- Learning \& Academic research institution for Master’s·PhD students, and Postdocs (G-LAMP) Program of the NRF funded by the Ministry of Education (RS-2025-25442252).

%
%

\clearpage
\bibliographystyle{splncs04}
\bibliography{main}

@String(NeurIPS = {Adv. Neural Inform. Process. Syst.})

@String(AAAI  = {AAAI})

@String(NeurIPS = {NeurIPS})

@article{loshchilov2017decoupled,
  title={Decoupled weight decay regularization},
  author={Loshchilov, Ilya and Hutter, Frank},
  journal={arXiv preprint arXiv:1711.05101},
  year={2017}
}

@article{dietterich1997solving,
  title={Solving the multiple instance problem with axis-parallel rectangles},
  author={Dietterich, Thomas G and Lathrop, Richard H and Lozano-P{\'e}rez, Tom{\'a}s},
  journal={Artificial intelligence},
  volume={89},
  number={1-2},
  pages={31--71},
  year={1997},
  publisher={Elsevier}
}

@article{wang2022internvideo,
  title={Internvideo: General video foundation models via generative and discriminative learning},
  author={Wang, Yi and Li, Kunchang and Li, Yizhuo and He, Yinan and Huang, Bingkun and Zhao, Zhiyu and Zhang, Hongjie and Xu, Jilan and Liu, Yi and Wang, Zun and others},
  journal={arXiv preprint arXiv:2212.03191},
  year={2022}
}

@inproceedings{devlin2019bert,
  title={Bert: Pre-training of deep bidirectional transformers for language understanding},
  author={Devlin, Jacob and Chang, Ming-Wei and Lee, Kenton and Toutanova, Kristina},
  booktitle={Proceedings of the 2019 conference of the North American chapter of the association for computational linguistics: human language technologies, volume 1 (long and short papers)},
  pages={4171--4186},
  year={2019}
}

@article{liu2019roberta,
  title={Roberta: A robustly optimized bert pretraining approach},
  author={Liu, Yinhan},
  journal={arXiv preprint arXiv:1907.11692},
  volume={364},
  year={2019}
}

@article{he2020deberta,
  title={Deberta: Decoding-enhanced bert with disentangled attention},
  author={He, Pengcheng and Liu, Xiaodong and Gao, Jianfeng and Chen, Weizhu},
  journal={arXiv preprint arXiv:2006.03654},
  year={2020}
}

@inproceedings{antol2015vqa,
  title={Vqa: Visual question answering},
  author={Antol, Stanislaw and Agrawal, Aishwarya and Lu, Jiasen and Mitchell, Margaret and Batra, Dhruv and Zitnick, C Lawrence and Parikh, Devi},
  booktitle={Proceedings of the IEEE international conference on computer vision},
  pages={2425--2433},
  year={2015}
}

@inproceedings{xiao2021next,
  title={Next-qa: Next phase of question-answering to explaining temporal actions},
  author={Xiao, Junbin and Shang, Xindi and Yao, Angela and Chua, Tat-Seng},
  booktitle={Proceedings of the IEEE/CVF conference on computer vision and pattern recognition},
  pages={9777--9786},
  year={2021}
}

@inproceedings{tapaswi2016movieqa,
  title={Movieqa: Understanding stories in movies through question-answering},
  author={Tapaswi, Makarand and Zhu, Yukun and Stiefelhagen, Rainer and Torralba, Antonio and Urtasun, Raquel and Fidler, Sanja},
  booktitle={Proceedings of the IEEE conference on computer vision and pattern recognition},
  pages={4631--4640},
  year={2016}
}

@inproceedings{jang2017tgif,
  title={Tgif-qa: Toward spatio-temporal reasoning in visual question answering},
  author={Jang, Yunseok and Song, Yale and Yu, Youngjae and Kim, Youngjin and Kim, Gunhee},
  booktitle={Proceedings of the IEEE conference on computer vision and pattern recognition},
  pages={2758--2766},
  year={2017}
}

@inproceedings{choi2021dramaqa,
  title={Dramaqa: Character-centered video story understanding with hierarchical qa},
  author={Choi, Seongho and On, Kyoung-Woon and Heo, Yu-Jung and Seo, Ahjeong and Jang, Youwon and Lee, Minsu and Zhang, Byoung-Tak},
  booktitle={Proceedings of the aaai conference on artificial intelligence},
  volume={35},
  number={2},
  pages={1166--1174},
  year={2021}
}

@inproceedings{min2024morevqa,
  title={Morevqa: Exploring modular reasoning models for video question answering},
  author={Min, Juhong and Buch, Shyamal and Nagrani, Arsha and Cho, Minsu and Schmid, Cordelia},
  booktitle={Proceedings of the IEEE/CVF Conference on Computer Vision and Pattern Recognition},
  pages={13235--13245},
  year={2024}
}

@article{yang2022zero,
  title={Zero-shot video question answering via frozen bidirectional language models},
  author={Yang, Antoine and Miech, Antoine and Sivic, Josef and Laptev, Ivan and Schmid, Cordelia},
  journal={Advances in Neural Information Processing Systems},
  volume={35},
  pages={124--141},
  year={2022}
}

@article{zhang2023video,
  title={Video-llama: An instruction-tuned audio-visual language model for video understanding},
  author={Zhang, Hang and Li, Xin and Bing, Lidong},
  journal={arXiv preprint arXiv:2306.02858},
  year={2023}
}

@inproceedings{li2022blip,
  title={Blip: Bootstrapping language-image pre-training for unified vision-language understanding and generation},
  author={Li, Junnan and Li, Dongxu and Xiong, Caiming and Hoi, Steven},
  booktitle={International conference on machine learning},
  pages={12888--12900},
  year={2022},
  organization={PMLR}
}

@article{dai2023instructblip,
  title={Instructblip: Towards general-purpose vision-language models with instruction tuning},
  author={Dai, Wenliang and Li, Junnan and Li, Dongxu and Tiong, Anthony and Zhao, Junqi and Wang, Weisheng and Li, Boyang and Fung, Pascale N and Hoi, Steven},
  journal={Advances in neural information processing systems},
  volume={36},
  pages={49250--49267},
  year={2023}
}

@article{liu2023visual,
  title={Visual instruction tuning},
  author={Liu, Haotian and Li, Chunyuan and Wu, Qingyang and Lee, Yong Jae},
  journal={Advances in neural information processing systems},
  volume={36},
  pages={34892--34916},
  year={2023}
}

@inproceedings{xiao2022video,
  title={Video graph transformer for video question answering},
  author={Xiao, Junbin and Zhou, Pan and Chua, Tat-Seng and Yan, Shuicheng},
  booktitle={European Conference on Computer Vision},
  pages={39--58},
  year={2022},
  organization={Springer}
}

@article{yin2019memory,
  title={Memory augmented deep recurrent neural network for video question answering},
  author={Yin, Chengxiang and Tang, Jian and Xu, Zhiyuan and Wang, Yanzhi},
  journal={IEEE transactions on neural networks and learning systems},
  volume={31},
  number={9},
  pages={3159--3167},
  year={2019},
  publisher={IEEE}
}

@article{sun2019learning,
  title={Learning video representations using contrastive bidirectional transformer},
  author={Sun, Chen and Baradel, Fabien and Murphy, Kevin and Schmid, Cordelia},
  journal={arXiv preprint arXiv:1906.05743},
  year={2019}
}

@inproceedings{radford2021learning,
  title={Learning transferable visual models from natural language supervision},
  author={Radford, Alec and Kim, Jong Wook and Hallacy, Chris and Ramesh, Aditya and Goh, Gabriel and Agarwal, Sandhini and Sastry, Girish and Askell, Amanda and Mishkin, Pamela and Clark, Jack and others},
  booktitle={International conference on machine learning},
  pages={8748--8763},
  year={2021},
  organization={PmLR}
}

@inproceedings{li2023blip,
  title={Blip-2: Bootstrapping language-image pre-training with frozen image encoders and large language models},
  author={Li, Junnan and Li, Dongxu and Savarese, Silvio and Hoi, Steven},
  booktitle={International conference on machine learning},
  pages={19730--19742},
  year={2023},
  organization={PMLR}
}

@inproceedings{xiao2024can,
  title={Can i trust your answer? visually grounded video question answering},
  author={Xiao, Junbin and Yao, Angela and Li, Yicong and Chua, Tat-Seng},
  booktitle={Proceedings of the IEEE/CVF Conference on Computer Vision and Pattern Recognition},
  pages={13204--13214},
  year={2024}
}

@inproceedings{liu2024timecraft,
  title={Timecraft: Navigate weakly-supervised temporal grounded video question answering via bi-directional reasoning},
  author={Liu, Huabin and Ma, Xiao and Zhong, Cheng and Zhang, Yang and Lin, Weiyao},
  booktitle={European Conference on Computer Vision},
  pages={92--107},
  year={2024},
  organization={Springer}
}

@inproceedings{xu2024exploring,
  title={Exploring Question Guidance and Answer Calibration for Visually Grounded Video Question Answering},
  author={Xu, Yuanxing and Wei, Yuting and Zhong, Shuai and Chen, Xinming and Qi, Jinsheng and Wu, Bin},
  booktitle={Findings of the Association for Computational Linguistics: EMNLP 2024},
  pages={3121--3133},
  year={2024}
}

@inproceedings{chen2025cross,
  title={Cross-modal causal relation alignment for video question grounding},
  author={Chen, Weixing and Liu, Yang and Chen, Binglin and Su, Jiandong and Zheng, Yongsen and Lin, Liang},
  booktitle={Proceedings of the Computer Vision and Pattern Recognition Conference},
  pages={24087--24096},
  year={2025}
}

@article{gupta2025toga,
  title={TOGA: Temporally Grounded Open-Ended Video QA with Weak Supervision},
  author={Gupta, Ayush and Roy, Anirban and Chellappa, Rama and Bastian, Nathaniel D and Velasquez, Alvaro and Jha, Susmit},
  journal={arXiv preprint arXiv:2506.09445},
  year={2025}
}

@article{wang2024grounded,
  title={Grounded-videollm: Sharpening fine-grained temporal grounding in video large language models},
  author={Wang, Haibo and Xu, Zhiyang and Cheng, Yu and Diao, Shizhe and Zhou, Yufan and Cao, Yixin and Wang, Qifan and Ge, Weifeng and Huang, Lifu},
  journal={arXiv preprint arXiv:2410.03290},
  year={2024}
}

@inproceedings{gao2019wslln,
  title={Wslln: Weakly supervised natural language localization networks},
  author={Gao, Mingfei and Davis, Larry and Socher, Richard and Xiong, Caiming},
  booktitle={Proceedings of the 2019 conference on empirical methods in natural language processing and the 9th international joint conference on natural language processing (EMNLP-IJCNLP)},
  pages={1481--1487},
  year={2019}
}

@inproceedings{huang2021cross,
  title={Cross-sentence temporal and semantic relations in video activity localisation},
  author={Huang, Jiabo and Liu, Yang and Gong, Shaogang and Jin, Hailin},
  booktitle={Proceedings of the IEEE/CVF international conference on computer vision},
  pages={7199--7208},
  year={2021}
}

@inproceedings{ma2020vlanet,
  title={Vlanet: Video-language alignment network for weakly-supervised video moment retrieval},
  author={Ma, Minuk and Yoon, Sunjae and Kim, Junyeong and Lee, Youngjoon and Kang, Sunghun and Yoo, Chang D},
  booktitle={European conference on computer vision},
  pages={156--171},
  year={2020},
  organization={Springer}
}

@inproceedings{mithun2019weakly,
  title={Weakly supervised video moment retrieval from text queries},
  author={Mithun, Niluthpol Chowdhury and Paul, Sujoy and Roy-Chowdhury, Amit K},
  booktitle={Proceedings of the IEEE/CVF Conference on Computer Vision and Pattern Recognition},
  pages={11592--11601},
  year={2019}
}

@article{alayrac2022flamingo,
  title={Flamingo: a visual language model for few-shot learning},
  author={Alayrac, Jean-Baptiste and Donahue, Jeff and Luc, Pauline and Miech, Antoine and Barr, Iain and Hasson, Yana and Lenc, Karel and Mensch, Arthur and Millican, Katherine and Reynolds, Malcolm and others},
  journal={Advances in neural information processing systems},
  volume={35},
  pages={23716--23736},
  year={2022}
}

@inproceedings{wu2021star_situated_reasoning,
author={Wu, Bo and Yu, Shoubin and Chen, Zhenfang and Tenenbaum, Joshua B and Gan, Chuang},
title = {{STAR}: A Benchmark for Situated Reasoning in Real-World Videos},
booktitle = {Thirty-fifth Conference on Neural Information Processing Systems (NeurIPS)},
year = {2021}
}

@inproceedings{miech2020end,
  title={End-to-end learning of visual representations from uncurated instructional videos},
  author={Miech, Antoine and Alayrac, Jean-Baptiste and Smaira, Lucas and Laptev, Ivan and Sivic, Josef and Zisserman, Andrew},
  booktitle={Proceedings of the IEEE/CVF conference on computer vision and pattern recognition},
  pages={9879--9889},
  year={2020}
}

@article{wang2021weakly,
  title={Weakly supervised temporal adjacent network for language grounding},
  author={Wang, Yuechen and Deng, Jiajun and Zhou, Wengang and Li, Houqiang},
  journal={IEEE Transactions on Multimedia},
  volume={24},
  pages={3276--3286},
  year={2021},
  publisher={IEEE}
}

\clearpage
\clearpage

\title{[Supplementary Material] \\
What You Ask is What You Ground:\\
Bridging Question Intent to Temporal Evidence\\ for Grounded VideoQA
}
\titlerunning{Supplementary Material}
\authorrunning{J. Seo et al.} 
\author{Jinhwan Seo\inst{1} \and
Kyubeom Han\inst{1} \and
Jumin Lee\inst{1} \and
Junhyug Noh\inst{2}\textsuperscript{$\dagger$} \and
Sung-eui Yoon\inst{1}\textsuperscript{$\dagger$}
}
\institute{KAIST \and Ewha Womans University\\
\email{\{jinhwan.seo,qbhan,jmlee\}@kaist.ac.kr, junhyug@ewha.ac.kr, sungeui@kaist.ac.kr}}

\maketitle

\appendix
\setcounter{equation}{12}
\setcounter{figure}{6}
\setcounter{table}{5}

\noindent
\begin{tabular}{@{}ll@{}}
\ref{sec:supp_piou} \hspace{1.2em} & \textbf{Analysis of Question-Invariant Grounding} \\
           & \hspace{1.2em} \ref{sec:supp_metric}\hspace{0.6em} Definitions of PIoU and GT-Pred Corr. \\
           & \hspace{1.2em} \ref{sec:supp_pious}\hspace{0.6em} PIoU Distribution \\
           & \hspace{1.2em} \ref{sec:supp_piou_bins}\hspace{0.6em} PIoU Robustness by Number of Questions \\
\ref{sec:supp_abl} \quad& \textbf{Additional Ablation Studies} \\
           & \hspace{1.2em} \ref{sec:supp_piou_abl}\hspace{0.6em} PIoU and GT--Pred Corr by Component \\
           & \hspace{1.2em} \ref{sec:supp_mil_alt}\hspace{0.6em} MIL Alternatives \\
           & \hspace{1.2em} \ref{sec:supp_qtype}\hspace{0.6em} Performance by Question Type\\
\ref{sec:supp_qual} \quad& \textbf{More Qualitative Results} \\
           & \hspace{1.2em} \ref{sec:supp_qual_success}\hspace{0.6em} Success Cases \\
           & \hspace{1.2em} \ref{sec:supp_similar_qns}\hspace{0.6em} Similar Questions \\
\ref{sec:supp_reproduce} & \textbf{Reproducibility} \\[2pt]
           & \hspace{1.2em} \ref{sec:supp_config}\hspace{0.6em} Implementation Configuration \\
           & \hspace{1.2em} \ref{sec:supp_cost}\hspace{0.6em} Training and Inference Cost \\
           & \hspace{1.2em} \ref{sec:supp_code}\hspace{0.6em} Code Release \\
\ref{sec:supp_limitation} & \textbf{Limitation} \\[2pt]
           & \hspace{1.2em} \ref{sec:supp_qual_failure}\hspace{0.6em} Failure Cases \\
           & \hspace{1.2em} \ref{sec:supp_discussion}\hspace{0.6em} Discussion \\
\end{tabular}

\setcounter{equation}{12}
\setcounter{figure}{6}
\setcounter{table}{5}

\section{Analysis of Question-Invariant Grounding}
\label{sec:supp_piou}
In the main paper, we identify \emph{question-invariant grounding} as a common failure mode in Grounded VideoQA, where models produce nearly identical temporal segments for different questions within the same video.
To quantify this behavior, we introduce two diagnostic metrics: Pairwise IoU (PIoU) and GT--Pred Correlation.
This section defines the two metrics (Sec.~\ref{sec:supp_metric}), visualizes PIoU distributions across methods (Sec.~\ref{sec:supp_pious}), and analyzes PIoU robustness with respect to the number of questions per video (Sec.~\ref{sec:supp_piou_bins}).

\subsection{Definitions of PIoU and GT--Pred Corr.}
\label{sec:supp_metric}

\noindent\textbf{Pairwise IoU (PIoU).}
PIoU measures whether a model produces question-conditioned grounding or collapses to a near-constant prediction per video.
For a video $v$ with questions $Q=\{q_1,\dots,q_n\}$, let $\hat{s}_i$ be the predicted temporal segment for $q_i$, and let $\mathcal{S}_i^{*}$ be the set of ground-truth segments for $q_i$.
For each question pair $(q_i,q_j)$, we first compute the ground-truth overlap as
\begin{equation}
  \mathrm{IoU}_{\mathrm{gt}}(q_i, q_j)
  = \max_{s \in \mathcal{S}_i^*,\, s' \in \mathcal{S}_j^*}
  \mathrm{IoU}(s, s'),
  \label{eq:iou_gt}
\end{equation}
where $\mathrm{IoU}(\cdot,\cdot)$ denotes temporal intersection-over-union. The predicted overlap is
$\mathrm{IoU}_{\mathrm{pred}}(q_i,q_j)=\mathrm{IoU}(\hat{s}_i,\hat{s}_j)$.

Questions that share the same evidence should legitimately overlap; including them would understate collapse.
We therefore retain only pairs whose ground-truth overlap is small:
\begin{equation}
  \mathcal{P}_v
  = \bigl\{(q_i,q_j)\mid i<j,\ \mathrm{IoU}_{\mathrm{gt}}(q_i,q_j)\le \tau \bigr\},
  \label{eq:piou_filter}
\end{equation}
where $\tau$ is a filtering threshold (default $\tau{=}0.5$).
PIoU is then the mean predicted overlap over all filtered pairs:
\begin{equation}
  \mathrm{PIoU}
  = \frac{1}{\bigl|\bigcup_v \mathcal{P}_v\bigr|}
    \sum_{v}\sum_{(q_i,q_j)\in\mathcal{P}_v}
  \mathrm{IoU}_{\mathrm{pred}}(q_i,q_j).
  \label{eq:piou}
\end{equation}
A question-invariant model yields $\mathrm{PIoU}\approx 1.0$, whereas a question-conditioned model produces low PIoU values, closer to the ground-truth overlap distribution.

\smallskip
\noindent\textbf{GT--Pred Correlation.}
PIoU captures whether predictions \emph{collapse}, but it does not tell whether a method preserves the \emph{relative overlap structure} implied by the ground truth.
To measure this, we consider all question pairs in each video without filtering,
$\mathcal{A}_v=\{(q_i,q_j)\mid i<j\}$,
and compute the Pearson correlation between ground-truth and predicted overlaps over
$\mathcal{A}=\bigcup_v \mathcal{A}_v$:
\begin{equation}
  r = \mathrm{Pearson}\!\Bigl(
    \{\mathrm{IoU}_{\mathrm{gt}}(q_i,q_j)\}_{(q_i,q_j)\in\mathcal{A}},\;
    \{\mathrm{IoU}_{\mathrm{pred}}(q_i,q_j)\}_{(q_i,q_j)\in\mathcal{A}}
  \Bigr).
  \label{eq:gt_pred_corr}
\end{equation}
A question-invariant method yields near-constant $\mathrm{IoU}_{\mathrm{pred}}$ across pairs, resulting in $r\approx 0$.
A question-conditioned method assigns higher predicted overlap to pairs whose ground-truth segments overlap and lower overlap to pairs that do not, producing $r>0$.
Together, PIoU and GT--Pred Correlation are complementary: PIoU detects collapse, while GT--Pred Correlation evaluates whether predicted overlaps vary \emph{in the correct direction}.

\subsection{PIoU Distribution Across Methods}
\label{sec:supp_pious}
\begin{figure}[t]
  \centering
  \includegraphics[width=0.9\linewidth]{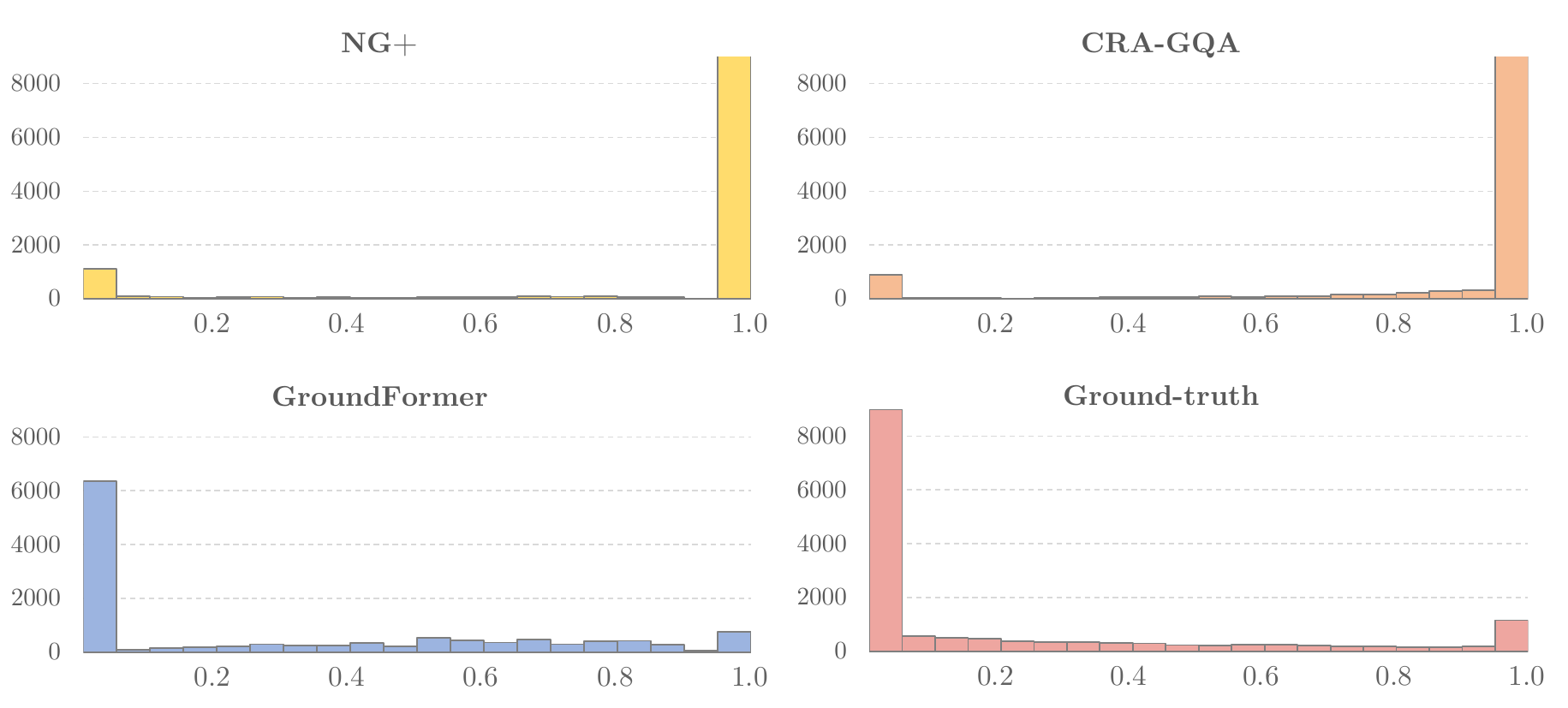}
  \vspace{-2mm}
  \caption{PIoU distribution on the NExT-GQA test split. Lower values indicate less question-invariant collapse.}
  \label{fig:supp_piou}
  \vspace{-5mm}
\end{figure}

\cref{fig:supp_piou} shows the PIoU histogram for each method on the NExT-GQA test split.
NG+~\cite{xiao2024can} and CRA-GQA~\cite{chen2025cross} place the majority of pairs near PIoU$=1.0$, indicating that different questions within the same video often receive nearly identical grounding.
In contrast, the ground-truth distribution concentrates near zero because many questions in a video target distinct, non-overlapping segments.
GroundFormer shifts the predicted distribution toward lower PIoU values, producing a profile that more closely resembles the ground truth.
This indicates that the gains reported in the main paper reflect broadly improved question-conditioned grounding rather than changes in a small number of outlier cases.

\subsection{PIoU Robustness by Number of Questions}
\label{sec:supp_piou_bins}

PIoU is defined over videos with at least two questions. In the NExT-GQA test split, 959 out of 990 videos (96.9\%) contain two or more questions and therefore contribute to PIoU; only 31 single-question videos are excluded by definition. The average number of questions per video is 5.6.

\cref{tab:piou_bins} shows that prior methods remain highly collapsed across all question-count bins, with PIoU consistently above 83. In contrast, GroundFormer remains close to the overall ground-truth PIoU of 21.3 across bins. This confirms that the PIoU improvement is not biased toward videos with a particular number of questions.

\begin{table}[t]
\caption{PIoU by the number of questions per video on the NExT-GQA test split. Lower is better.}
\label{tab:piou_bins}
\centering
\setlength{\tabcolsep}{5pt}
\renewcommand{\arraystretch}{1.05}
\resizebox{0.82\linewidth}{!}{
\begin{tabular}{lcccc}
\toprule
Method & 2Q (8.6\%) & 3Q (9.5\%) & 4--5Q (28.3\%) & 6+Q (50.5\%) \\
\midrule
NG+~\cite{xiao2024can}       & 83.2 & 89.9 & 87.7 & 87.9 \\
CRA-GQA~\cite{chen2025cross} & 88.3 & 87.6 & 88.2 & 88.9 \\
\textbf{GroundFormer}        & \textbf{23.7} & \textbf{22.0} & \textbf{28.7} & \textbf{29.6} \\
\bottomrule
\end{tabular}}
\end{table}


\section{Additional Ablation Studies}
\label{sec:supp_abl}
This section provides additional ablations beyond the main paper.
We first analyze how structural design choices and auxiliary components affect question-invariant grounding using PIoU and GT--Pred Correlation (Sec.~\ref{sec:supp_piou_abl}).
We then compare MIL Cross-Attention with alternative weakly supervised grounding objectives (Sec.~\ref{sec:supp_mil_alt}) and report performance by question type on NExT-GQA and STAR (Sec.~\ref{sec:supp_qtype}).

\begin{table}[t!]
\caption{Ablation on NExT-GQA with question-invariant grounding metrics.
PIoU: lower is better (less collapse). GT--Pred Corr.: higher is better (better agreement with the ground-truth overlap structure).}
\label{tab:ablation_piou}
\centering
\setlength{\tabcolsep}{3.6pt}
\renewcommand{\arraystretch}{1.0}
\resizebox{\textwidth}{!}{
\begin{tabular}{cc|cccc|cc|cc|cc|cc}
\toprule
\multicolumn{2}{c|}{\textbf{Structure}} &
\multicolumn{4}{c|}{\textbf{Aux.\ Comp.}} &
\multicolumn{2}{c|}{\textbf{QA}} &
\multicolumn{2}{c|}{\textbf{IoP}} &
\multicolumn{2}{c|}{\textbf{IoU}} &
\multicolumn{2}{c}{\textbf{Question-Invariant}} \\
\cmidrule(lr){1-2}\cmidrule(lr){3-6}\cmidrule(lr){7-8}\cmidrule(lr){9-10}\cmidrule(lr){11-12}\cmidrule(lr){13-14}
\makecell{Two-pass\\train} &
\makecell{Comm.\\tokens} &
$\mathcal{L}_{type}$ & $\mathcal{L}_{mcl}$ & $\mathcal{G}_{\sigma}$ & MIL &
\textbf{Acc@GQA} & Acc@VQA &
mIoP & TIoP@0.5 &
mIoU & TIoU@0.5 &
\textbf{PIoU} $\downarrow$ & \makecell{\textbf{GT--Pred Corr.} $\uparrow$} \\
\midrule\midrule
\cmark &  &  &  &  &  &
14.8 & 58.5 & 26.5 & 24.8 & 14.7 & 10.6 & 57.4 & 0.06 \\
\cmark & \cmark &  &  &  &  &
18.6 & 59.8 & 29.6 & 28.6 & 17.0 & 13.0 & 43.5 & 0.10 \\
\cmark & \cmark & \cmark &  &  &  &
19.0 & 60.4 & 30.5 & 30.2 & 16.9 & 13.2 & 37.9 & 0.10 \\
\cmark & \cmark & \cmark & \cmark &  &  &
20.1 & 61.2 & 32.0 & 30.3 & 17.2 & 12.5 & 29.4 & \textbf{0.12} \\
\cmark & \cmark & \cmark & \cmark & \cmark &  &
20.8 & 61.2 & 32.4 & 32.0 & 16.5 & 12.0 & 29.4 & \textbf{0.12} \\
\midrule
& \cmark &  &  & \cmark & \cmark &
15.7 & 58.2 & 26.7 & 25.9 & 14.6 & 10.8 & 51.4 & 0.05 \\
\cmark & \cmark & \cmark & \cmark & \cmark & \cmark &
\textbf{21.5} & \textbf{61.7} & \textbf{34.0} & \textbf{33.7} &
\textbf{17.5} & \textbf{12.8} & \textbf{28.2} & \textbf{0.12} \\
\bottomrule
\end{tabular}}
\end{table}

\subsection{PIoU and GT--Pred Corr.\ by Component}
\label{sec:supp_piou_abl}
\cref{tab:ablation_piou} shows that reducing question-invariant grounding requires both \emph{structural conditioning} and \emph{appropriate supervision}.
Starting from the two-pass baseline (row~1), PIoU remains high (57.4), indicating substantial collapse across questions.
Adding communication tokens (row~2) sharply lowers PIoU (43.5) and increases GT--Pred Corr.\ (0.06$\rightarrow$0.10), confirming that early visuo-lingual conditioning is central to question-specific grounding.
The auxiliary losses further improve question sensitivity: $\mathcal{L}_{type}$ (row~3) reduces PIoU to 37.9, and adding $\mathcal{L}_{mcl}$ (row~4) yields the largest drop to 29.4 with the strongest correlation (0.12), suggesting that type priors and hierarchical alignment provide complementary supervision for preserving question-dependent overlap structure.

In contrast, Gaussian smoothing alone (row~5) leaves PIoU and GT--Pred Corr.\ unchanged, consistent with its role as a temporal \emph{refinement} step rather than a source of question conditioning.
Finally, enabling MIL cross-attention together with two-pass training and auxiliary objectives (row~7) achieves the best overall behavior, with the lowest PIoU (28.2) and highest GT--Pred Corr.\ (0.12), while also improving Acc@GQA and localization metrics.
Notably, MIL without two-pass training (row~6) improves neither PIoU nor GT--Pred Corr.\ (51.4 / 0.05), indicating that MIL factorization alone is insufficient without the training-time question pass that promotes question-discriminative visual representations.

\subsection{Comparison with Weakly Supervised Grounding Alternatives}
\label{sec:supp_mil_alt}

To examine whether the gains come from the proposed candidate--temporal MIL factorization or simply from adding a weakly supervised temporal grounding objective, we compare MIL Cross-Attention with three weakly supervised moment-localization objectives: TGA~\cite{mithun2019weakly}, MIL-NCE~\cite{miech2020end}, and WSTAN~\cite{wang2021weakly}.
All variants use the same backbone and apply Gaussian smoothing for fair comparison.

\begin{table}[t]
\caption{Comparison with weakly supervised grounding alternatives on NExT-GQA.}
\label{tab:mil_alternatives}
\centering
\setlength{\tabcolsep}{6pt}
\renewcommand{\arraystretch}{1.05}
\resizebox{0.72\linewidth}{!}{
\begin{tabular}{lcccc}
\toprule
Method & \textbf{Acc@GQA} & Acc@VQA & mIoP & mIoU \\
\midrule
Cross attention & 14.8 & 58.5 & 26.5 & 14.7 \\
TGA~\cite{mithun2019weakly}       & 19.5 & 59.5 & 31.6 & 17.2 \\
MIL-NCE~\cite{miech2020end}       & 19.1 & 59.4 & 31.4 & 16.7 \\
WSTAN~\cite{wang2021weakly}       & 20.0 & 60.9 & 31.9 & \textbf{18.3} \\
\textbf{GroundFormer}             & \textbf{21.5} & \textbf{61.7} & \textbf{34.0} & 17.5 \\
\bottomrule
\end{tabular}}
\end{table}

As shown in \cref{tab:mil_alternatives}, existing weakly supervised moment-localization objectives improve over standard cross-attention, confirming that temporal supervision signals are useful for Grounded VideoQA.
However, they mainly optimize temporal selection independently of answer-candidate competition.
GroundFormer performs best on Acc@GQA, Acc@VQA, and mIoP, supporting our design choice that temporal evidence should be selected jointly with answer candidates.

\begin{table}[t!]
\caption{Per-type performance on NExT-GQA~\cite{xiao2024can}.
}
\label{tab:supp_qtype}
\centering
\setlength{\tabcolsep}{5.2pt}
\renewcommand{\arraystretch}{1.0}
\resizebox{0.85\linewidth}{!}{
\begin{tabular}{l|cc|cc|cc}
\toprule
Method & \multicolumn{2}{c|}{Question Type} & \textbf{Acc@GQA} & Acc@VQA & TIoP@0.5 & TIoU@0.5 \\
\midrule\midrule
\multirow{5}{*}{NG+~\cite{xiao2024can}}
& \multirow{2}{*}{Causal}   & Why      & 16.9 & 60.1 & 27.6 & 9.3 \\
&                           & How      & 17.2 & \textbf{60.3} & 27.6 & 9.4 \\
\cmidrule(lr){2-7}
& \multirow{3}{*}{Temporal} & Before\&After & 12.0 & 59.3 & 19.5 & 9.0 \\
&                           & When     & 17.5 & \textbf{63.0} & 27.0 & 9.4 \\
&                           & Present  & 10.8 & 60.3 & 19.4 & 4.3 \\
\midrule
\multirow{5}{*}{GroundFormer}
& \multirow{2}{*}{Causal}   & Why      & \textbf{22.2} & \textbf{61.9} & \textbf{35.6} & \textbf{13.1} \\
&                           & How      & \textbf{23.1} & \textbf{60.3} & \textbf{37.1} & \textbf{12.7} \\
\cmidrule(lr){2-7}
& \multirow{3}{*}{Temporal} & Before\&After & \textbf{17.3} & \textbf{59.6} & \textbf{26.6} & \textbf{12.3} \\
&                           & When     & \textbf{25.8} & 60.3 & \textbf{37.7} & \textbf{13.3} \\
&                           & Present  & \textbf{12.9} & \textbf{61.3} & \textbf{23.7} & \textbf{8.6} \\
\bottomrule
\end{tabular}}
\vspace{-5mm}
\end{table}

\begin{table}[t!]
\caption{Per-type performance on STAR~\cite{wu2021star_situated_reasoning}. 
}
\label{tab:supp_qtype_star}
\centering
\setlength{\tabcolsep}{5.6pt}
\renewcommand{\arraystretch}{1.0}
\resizebox{0.85\linewidth}{!}{
\begin{tabular}{l|c|cc|cc}
\toprule
Method & \textbf{Question Type} & \textbf{Acc@GQA} & Acc@VQA & TIoP@0.5 & TIoU@0.5 \\
\midrule\midrule
\multirow{4}{*}{NG+~\cite{xiao2024can}}
& Interaction & 12.5 & 52.3 & 23.6 & 5.6 \\
& Sequence    & 32.0 & 59.5 & 53.2 & 4.2 \\
& Prediction  & 33.2 & 61.5 & 51.9 & 5.3 \\
& Feasibility & 16.1 & 61.5 & \textbf{28.4} & 4.2 \\
\midrule
\multirow{4}{*}{CRA-GQA~\cite{chen2025cross}}
& Interaction & 14.1 & 53.3 & 25.5 & 7.5 \\
& Sequence    & 34.9 & 60.9 & 57.3 & 3.8 \\
& Prediction  & \textbf{36.5} & 62.8 & \textbf{56.9} & 4.6 \\
& Feasibility & \textbf{17.6} & 63.3 & 27.8 & 8.6 \\
\midrule
\multirow{4}{*}{\textbf{GroundFormer}}
& Interaction & \textbf{22.4} & \textbf{56.0} & \textbf{38.2} & \textbf{24.1} \\
& Sequence    & \textbf{37.8} & \textbf{63.8} & \textbf{58.1} & \textbf{10.8} \\
& Prediction  & 25.5 & \textbf{64.7} & 38.0 & \textbf{5.4} \\
& Feasibility & 13.9 & \textbf{65.1} & 23.1 & \textbf{11.0} \\
\bottomrule
\end{tabular}}
\end{table}

\subsection{Performance by Question Type}
\label{sec:supp_qtype}
\cref{tab:supp_qtype,tab:supp_qtype_star} report per-type results on NExT-GQA and STAR, respectively.
On NExT-GQA, GroundFormer consistently improves Acc@GQA and localization at IoP@0.5 (TIoP@0.5) over NG+ across all causal and temporal categories, with the largest gain on \textit{Temporal--When}, which requires pinpointing a specific moment.
On STAR, GroundFormer improves Acc@GQA across \textit{Interaction} and \textit{Sequence} questions with substantial gains in temporal IoU at 0.5, indicating more faithful segment boundaries.
Performance is weaker on \textit{Prediction} questions in terms of IoU, suggesting that forward-looking reasoning remains challenging when the supporting evidence is subtle or distributed.


\section{More Qualitative Results}
\label{sec:supp_qual}
We provide additional qualitative results on NExT-GQA to illustrate how GroundFormer grounds temporal evidence in a question-dependent manner.
\cref{sec:supp_qual_success} presents success cases where distinct questions within the same video are localized to distinct temporal segments.
\cref{sec:supp_similar_qns} further examines whether the model assigns similar intervals to semantically related questions while separating questions that target different events.

\begin{figure}[b!]
  \vspace{-0.3cm}
  \centering
  \includegraphics[width=0.95\linewidth]{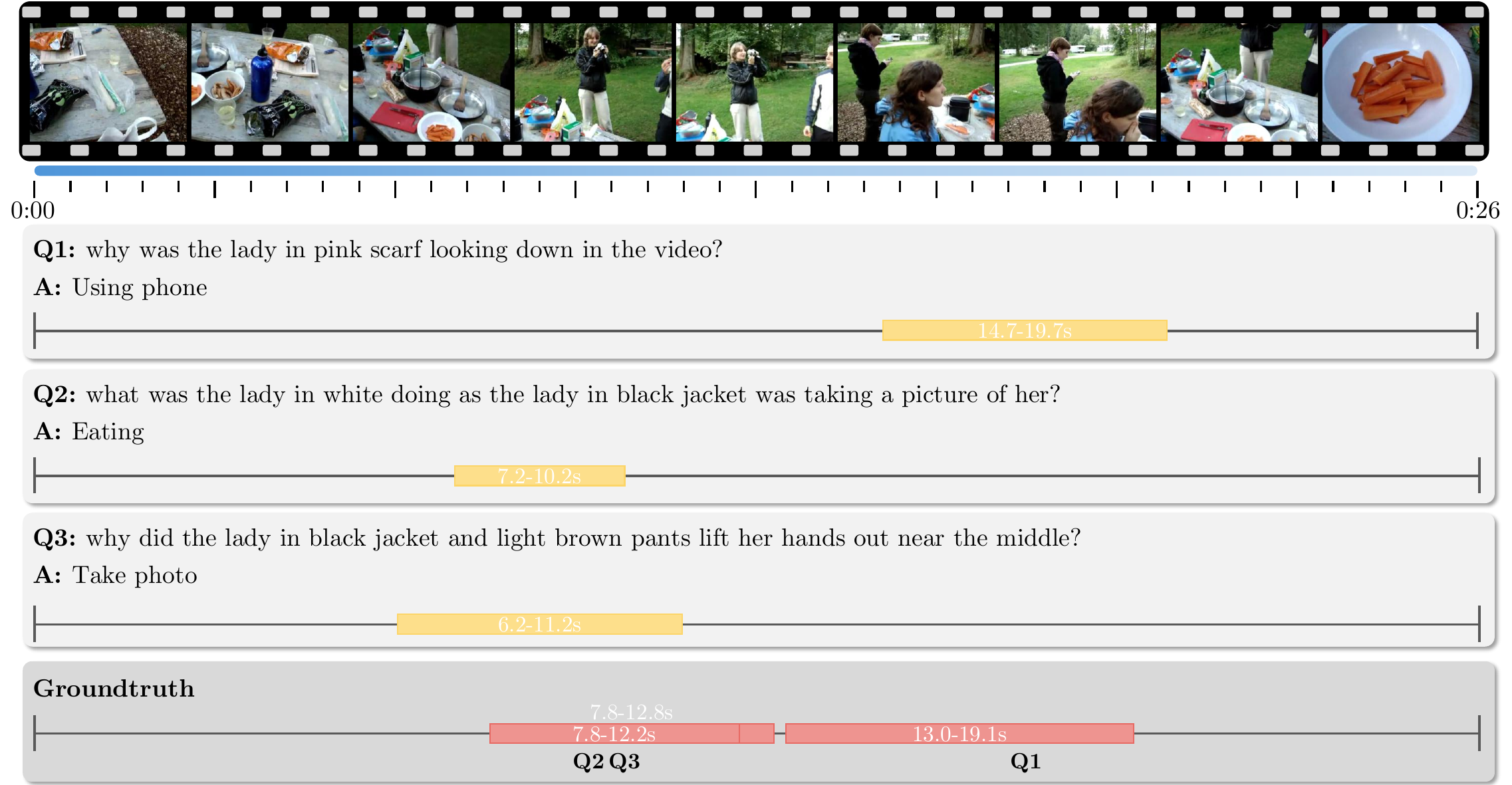}
  \caption{Additional qualitative result on NExT-GQA. GroundFormer produces distinct, question-specific grounding intervals within the same video.}
  \label{fig:supp_qual1}
  \vspace{-0.2cm}
\end{figure}

\begin{figure}[t!]
  \centering
  \includegraphics[width=0.95\linewidth]{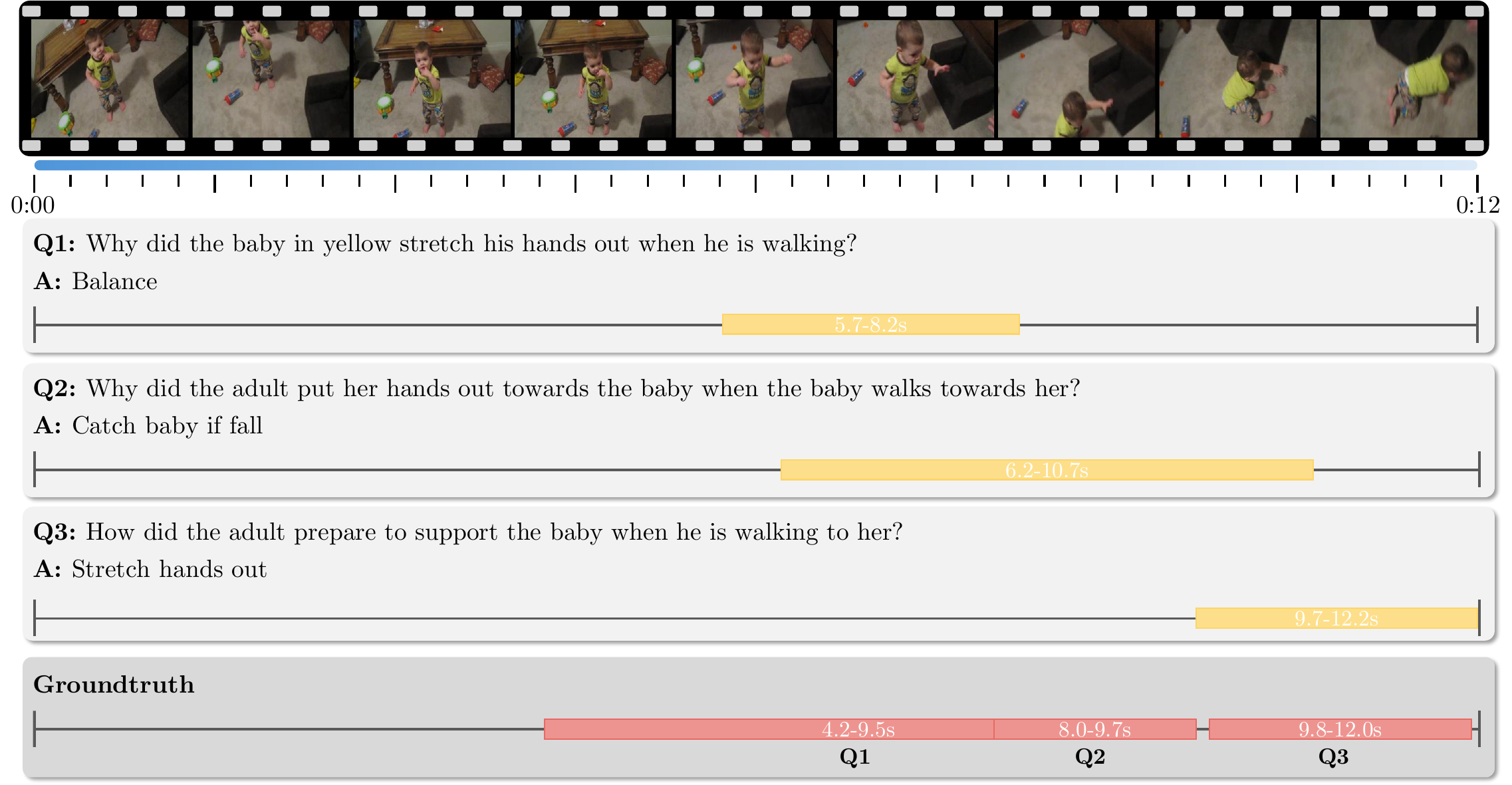}
  \caption{Additional qualitative result on NExT-GQA. GroundFormer preserves temporal ordering across questions and avoids collapsing multiple questions to a single segment.}
  \label{fig:supp_qual2}
  \vspace{-0.2cm}
\end{figure}

\subsection{Success Cases}
\label{sec:supp_qual_success}

\cref{fig:supp_qual1,fig:supp_qual2} show representative success cases.
In \cref{fig:supp_qual1}, three questions refer to different people and moments within a 26-second video; GroundFormer localizes each question to a different interval that closely matches the ground-truth.
In \cref{fig:supp_qual2}, the questions target successive stages of an interaction (walking, reaching out, preparing support) in a 12-second clip, and the predicted segments follow the ground-truth temporal ordering without collapsing to a single interval.

\begin{figure}[b!]
  \vspace{-0.3cm}
  \centering
  \includegraphics[width=0.99\linewidth]{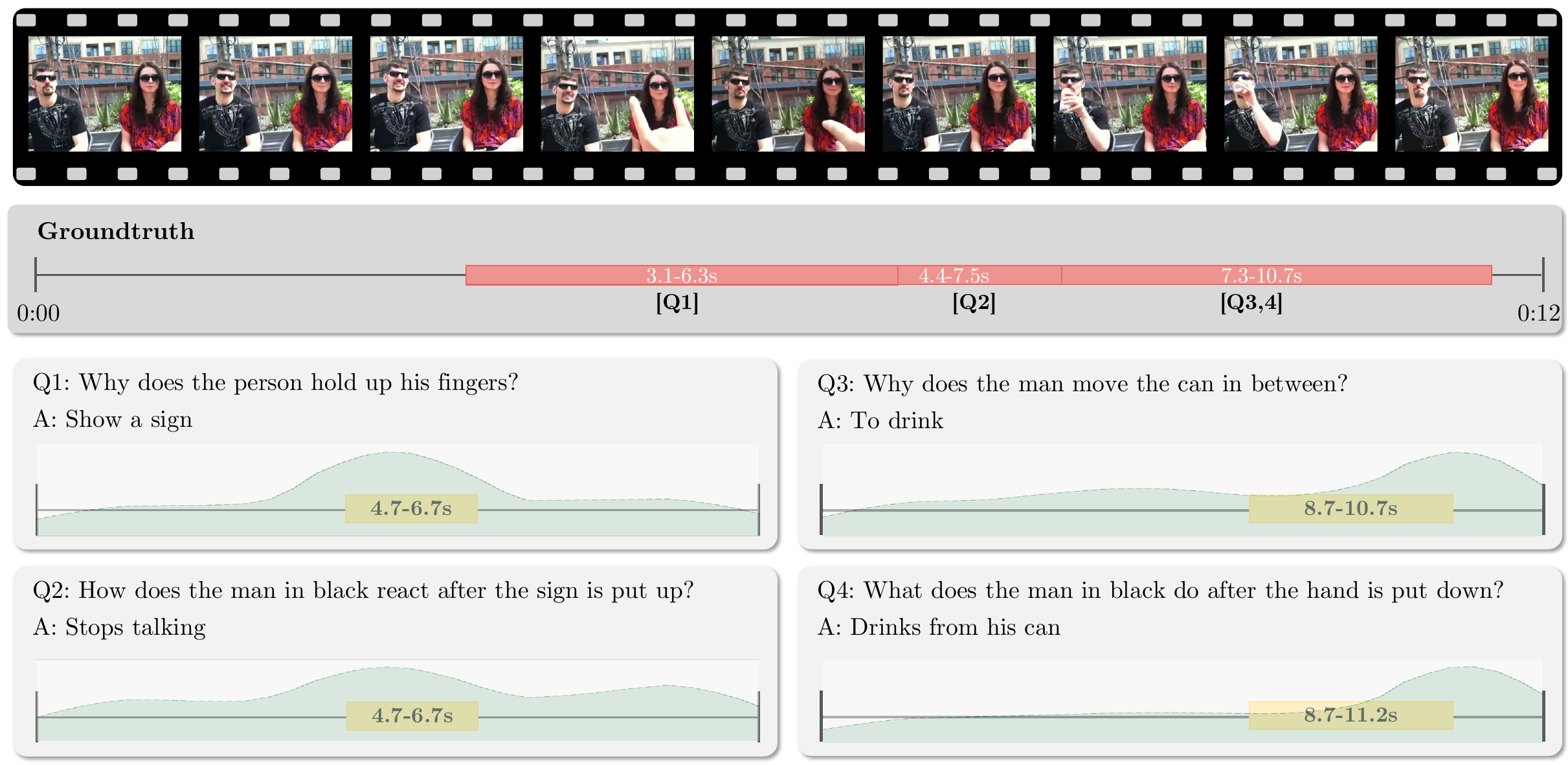}
  \caption{Grounding consistency for semantically related questions within a video: questions about the same event receive overlapping intervals, while different events are separated.}
  \label{fig:supp_qual_sim}
  \vspace{-0.3cm}
\end{figure}

\begin{figure}[t!]
  \vspace{-0.3cm}
  \centering
  \includegraphics[width=0.99\linewidth]{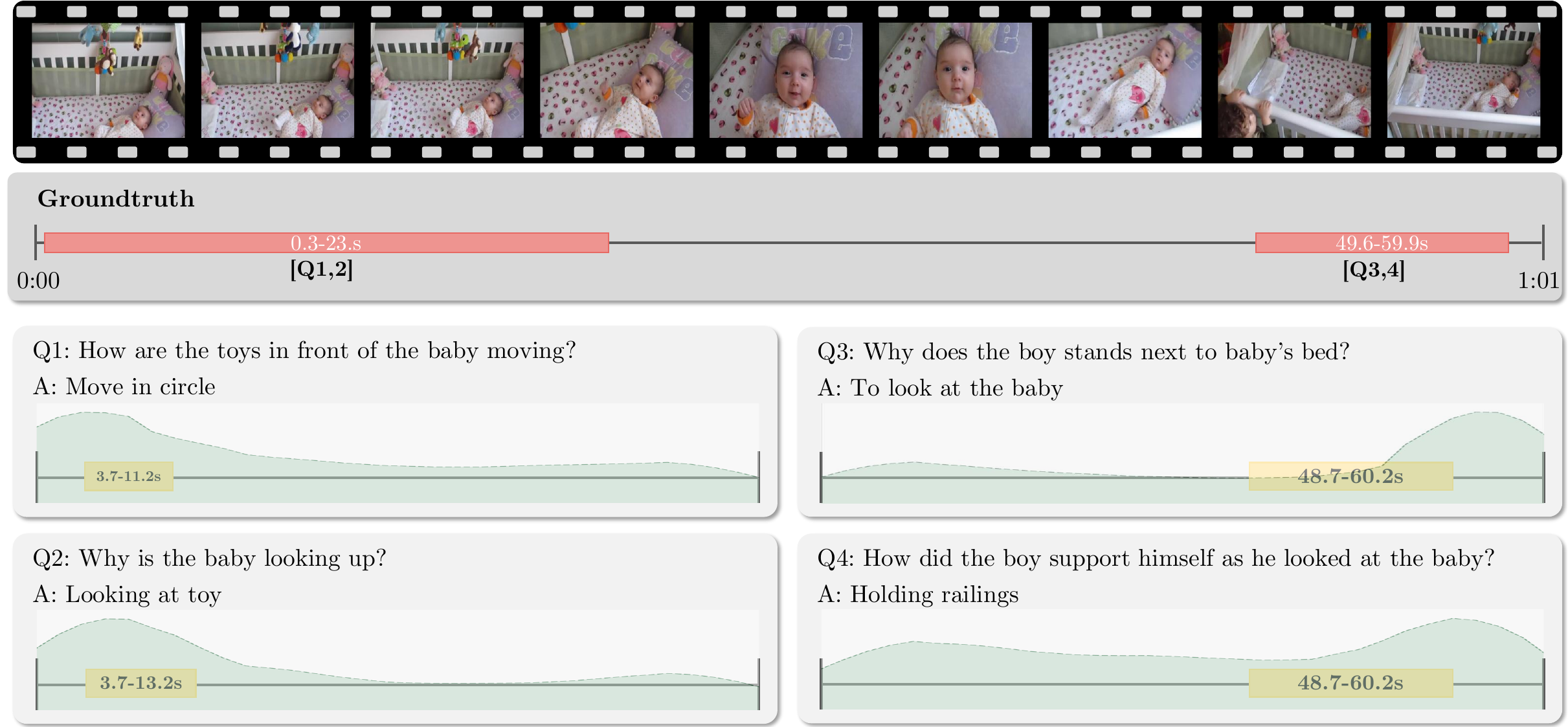}
  \caption{Grounding consistency on a longer video: early-event questions (baby/toys) and late-event questions (boy/crib) are localized to distinct temporal regions.}
  \label{fig:supp_qual_sim2}
  \vspace{-0.2cm}
\end{figure}

\subsection{Similar Questions}
\label{sec:supp_similar_qns}
\cref{fig:supp_qual_sim,fig:supp_qual_sim2} examine whether GroundFormer yields \emph{consistent} grounding for semantically related questions while remaining \emph{discriminative} across distinct events within the same video.
In \cref{fig:supp_qual_sim}, Q1 and Q2 both refer to the sign-related event and are grounded to the same interval (4.7--6.7s), whereas Q3 and Q4 both target the later drinking action and are localized to a different interval (8.7--10.7s); importantly, the two event groups are clearly separated.
\cref{fig:supp_qual_sim2} shows the same behavior in a longer 61-second video: Q1--Q2 about the baby and toys map to an early segment (3.7--13.2s), while Q3--Q4 about the boy near the crib concentrate on a late segment (48.7--60.2s).
Overall, GroundFormer assigns overlapping intervals when questions share the same temporal evidence and produces distinct segments when questions refer to different events, indicating that grounding is driven by question semantics rather than superficial phrasing.

\section{Reproducibility}
\label{sec:supp_reproduce}
This section provides details needed to reproduce our experiments.
We summarize the implementation configuration and hyperparameters in Sec.~\ref{sec:supp_config}, report training and inference cost in Sec.~\ref{sec:supp_cost}, and describe the code release plan in Sec.~\ref{sec:supp_code}.

\subsection{Implementation Configuration}
\label{sec:supp_config}
\cref{tab:config} summarizes the model architecture, training setup, and key hyperparameters used for all experiments. Unless otherwise stated, we use the same configuration for both benchmarks.

\begin{table}[h]
\vspace{-0.5cm}
\centering
\caption{Implementation configuration and hyperparameters.
}
\label{tab:config}
\setlength{\tabcolsep}{7pt}
\renewcommand{\arraystretch}{1.05}
\resizebox{0.75\textwidth}{!}{
\begin{tabular}{@{}l@{\hskip 10pt}l@{\hskip 10pt}l@{}}
\toprule
\textbf{Component} & \textbf{Hyperparameter} & \textbf{Value} \\
\midrule
\multirow{5}{*}{\textit{Encoders}}
& Vision encoder            & CLIP ViT-L/14 \\
& Vision dim.\ ($D_V$)      & 768 \\
& Language encoder          & RoBERTa-base \\
& Language dim.\ ($D_L$)    & 768 \\
& Projection dim.\ ($D$)    & 768 \\
\midrule
\multirow{2}{*}{\textit{Video sampling}}
& Frames ($T$)              & 32 (uniform) \\
& Frame resolution          & $224\times224$ \\
\midrule
\textit{Comm.\ tokens}
& Query tokens ($N$)        & 32 \\
\midrule
\multirow{4}{*}{\textit{GroundFormer}}
& Transformer layers        & 4 \\
& Attention heads           & 12 \\
& FFN hidden dim.           & 3072 \\
& Gaussian kernel ($\sigma$)& 2 \\
\midrule
\multirow{2}{*}{\textit{Two-pass}}
& Question candidates ($Q_n$) & 5 (1\,GT + 4\,neg.) \\
& Answer candidates ($A_n$)   & NExT-GQA: 5,\quad STAR: 4 \\
\midrule
\multirow{7}{*}{\textit{Training}}
& Optimizer                 & AdamW \\
& Learning rate             & $1\times10^{-5}$ \\
& Weight decay              & $1\times10^{-2}$ \\
& LR schedule               & Cosine decay \\
& Epochs                    & 30 \\
& Batch size                & 16 \\
& Temperature ($\tau$)      & 0.07 \\
\bottomrule
\end{tabular}}
\vspace{-0.3cm}
\end{table}

\subsection{Training and Inference Cost}
\label{sec:supp_cost}

We measure per-epoch training time on an RTX 3090 GPU.
GroundFormer takes approximately $2.1\times$ the per-epoch training time of NG+, mainly because the training-time question pass is added to enable hierarchical V$\rightarrow$Q$\rightarrow$A contrastive alignment.
At inference, however, GroundFormer executes only the answer pass.
Thus, the two-pass design increases training cost but introduces no additional inference pass.

\subsection{Code Release}
\label{sec:supp_code}
We provide the full codebase as a \textbf{zip file} in the supplementary material and will \textbf{publicly release} the repository with pre-trained checkpoints upon acceptance.

\section{Limitations}
\label{sec:supp_limitation}

\subsection{Failure Cases}
\label{sec:supp_qual_failure}
Following ECCV's suggested author practices, we summarize representative limitations observed in qualitative analysis and clarify the conditions under which our current formulation can fail.

\begin{figure}[t!]
  \centering
    \includegraphics[width=0.9\linewidth]{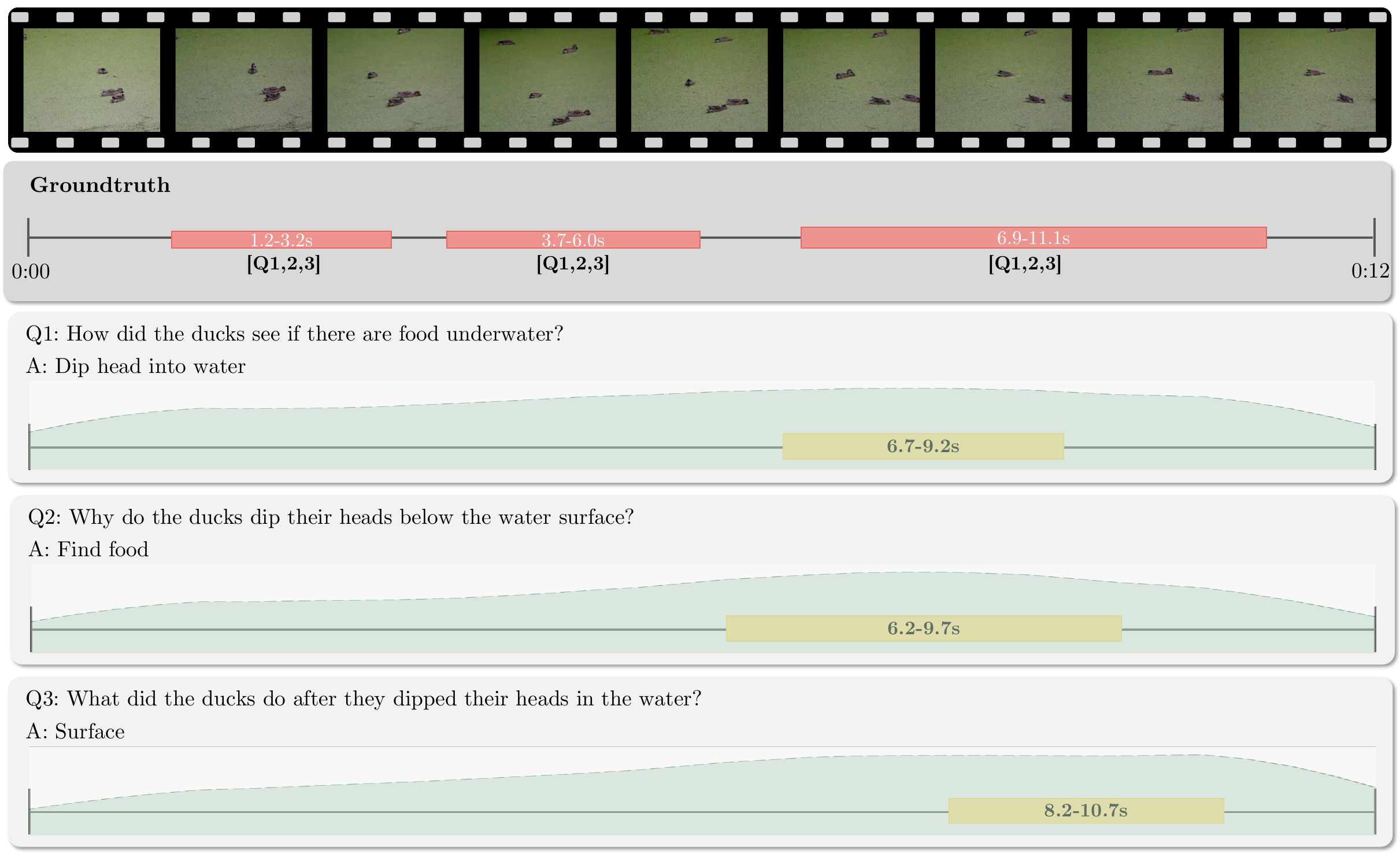}
  \caption{Failure case: repeated actions with evidence appearing in multiple disjoint intervals.}
  \label{fig:supp_fail}
\end{figure}

\smallskip
\noindent\textbf{Repeated actions and multi-interval evidence.}
In \cref{fig:supp_fail}, all three questions describe repeated duck-feeding behavior (\emph{e.g.}, dipping, submerging, surfacing). Because the relevant evidence recurs across the video, the ground-truth spans multiple disjoint temporal positions. Our model predicts only a single interval and thus localizes only the late occurrence, missing earlier repetitions. This exposes a limitation of our \emph{single-interval} grounding formulation: when evidence appears multiple times, one predicted segment per question cannot capture all relevant instances.

\begin{figure}[t!]
  \centering
    \includegraphics[width=0.9\linewidth]{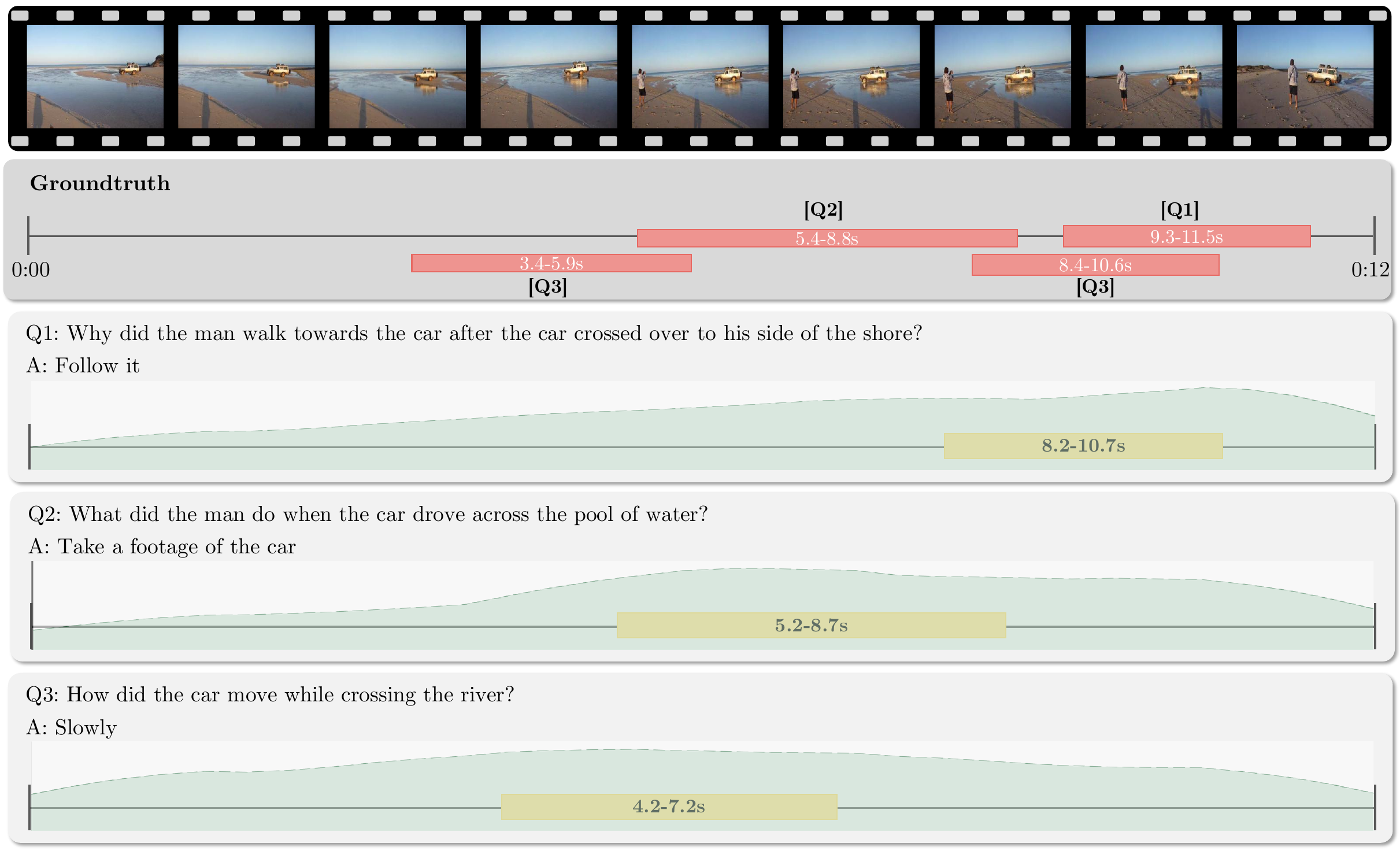}
  \caption{Failure case: fine-grained motion cues (\emph{e.g.}, deceleration) under sparse temporal sampling.}
  \label{fig:supp_fail2}
\end{figure}

\smallskip
\noindent\textbf{Fine-grained motion.}
\cref{fig:supp_fail2} shows a car crossing a river while a man films from the shore. Q1 and Q2 are grounded correctly, but Q3 has two ground-truth deceleration segments (3.4--5.9s and 8.4--10.6s), and the model captures only the earlier one. This failure reflects two compounding factors: (i) single-interval grounding cannot represent repeated evidence, and (ii) uniform sampling at $T{=}32$ frames may be temporally too coarse to reliably capture subtle speed changes. Distinguishing deceleration from ordinary crossing often requires fine-grained motion cues that can be attenuated by sparse sampling and by using frozen frame-level features.

\subsection{Discussion}
\label{sec:supp_discussion}
This work identifies \textbf{\emph{question-invariant grounding}} as a failure mode in weakly-supervised Grounded VideoQA and shows that early visuo-lingual conditioning can substantially mitigate question-invariant grounding. However, question-conditioned grounding is not sufficient for faithful localization in all cases. In particular, our current design predicts a \emph{single} temporal segment per question, which is mismatched to scenarios where evidence repeats across disjoint intervals. Moreover, the common practice of uniform sparse sampling can discard fine-grained temporal cues needed for subtle action distinctions. These limitations suggest promising directions such as multi-interval grounding formulations and more temporally sensitive representations. We also expect the diagnostic metrics introduced in this work (PIoU and GT--Pred Correlation) to complement standard localization scores by explicitly measuring whether grounding behavior varies with question semantics.






\end{document}